%% file: main.tex
\documentclass[twoside]{article}
\usepackage[accepted]{aistats2023}
\usepackage{soul,color}
\usepackage{aistats2023}
\usepackage{url} %fixes linebreaks in reference urls!
\usepackage{breqn} %used to fix the split of equations into a better way.
\usepackage{natbib}
\usepackage{hyperref}

\setcitestyle{numbers,sort,round}
\setcitestyle{authoryear,open={(},close={)}, aysep={,}} 
\usepackage{bm}
\usepackage{booktabs}
\usepackage{lscape}
\usepackage{apalike}
\usepackage{amsmath}
\usepackage{amsfonts}
\usepackage{graphicx}
\usepackage{subcaption}

\begin{document}


\runningauthor{Nakis, \c{C}elikkanat, Boucherie, Burmester, Djurhuus, Holmelund, Frolcová, Mørup}

\twocolumn[

\aistatstitle{Characterizing Polarization in Social Networks using the Signed Relational Latent Distance Model}

\aistatsauthor{ 
Nikolaos Nakis \And Abdulkadir \c{C}elikkanat \And  Louis Boucherie \And Christian Djurhuus}
\aistatsaddress{Technical University \\of Denmark \And  Technical University \\of Denmark \And Technical University \\of Denmark \And Technical University \\of Denmark}

\aistatsauthor{ 
Felix Burmester \And Daniel Mathias Holmelund \And Monika Frolcová \And Morten Mørup }
\aistatsaddress{Technical University \\of Denmark \And  Technical University \\of Denmark \And Technical University \\of Denmark \And Technical University \\of Denmark}]

\begin{abstract}
    Graph representation learning has become a prominent tool for the characterization and understanding of the structure of networks in general and social networks in particular. Typically, these representation learning approaches embed the networks into a low-dimensional space in which the role of each individual can be characterized in terms of their latent position. A major current concern in social networks is the emergence of polarization and filter bubbles promoting a mindset of ''us-versus-them'' that may be defined by extreme positions believed to ultimately lead to political violence and the erosion of democracy. Such polarized networks are typically characterized in terms of signed links reflecting likes and dislikes. We propose the Signed Latent Distance Model (\textsc{SLDM}) utilizing for the first time the Skellam distribution as a likelihood function for signed networks. We further extend the modeling to the characterization of distinct extreme positions by constraining the embedding space to polytopes, forming the \textsc{S}igned \textsc{L}atent relational d\textsc{I}stance \textsc{M}odel (\textsc{SLIM}). On four real social signed networks of polarization, we demonstrate that the models extract low-dimensional characterizations that well predict friendships and animosity while \textsc{SLIM} provides interpretable visualizations defined by extreme positions when restricting the embedding space to polytopes.
\end{abstract}

\input{1-introduction.tex}

\input{3-proposed_model.tex}

\input{4-experiments.tex}

\input{5-conclusion.tex}
\section*{Acknowledgements}
We would like to express sincere appreciation and thank the reviewers for their constructive feedback and their insightful comments. We gratefully acknowledge the Independent Research Fund Denmark for supporting this work [grant number: 0136-00315B].

\bibliography{reference}
\appendix
\input{supplement.tex}

\end{document}

%% file: 1-introduction.tex
\section{INTRODUCTION}\label{sec:introduction}

For several decades, the origin and influence of political polarization have been issues receiving considerable attention both within scholarly research and the public media \citep{hetherington_2009}. 
% We need here to provide an overview of existing network science literature on polarization
Several studies have demonstrated an increasing partisan polarization among the political elites, some of which rely on network science approaches, for instance, using co-voting similarity networks and modularity to model and explain the distinct aspects of the data \citep{moody_mucha_2013}.
Whereas polarization has been described in terms of communities and their boundary properties \citep{guerra2013measure}, latent distance modeling has also been used to extract bipolar structures %to be examined 
\citep{barbera2015tweeting}.

%We presently \hl{(Who are we in this context?)} focus on polarization as expressed in terms of extremes defined by an "us-versus-them" mentality \cite{dagnes2019us}.

%Might fit in here? But could be excluded entirely.
%The conceptual idea of polytopes as formed by pure types can be traced back to Plato's forms, which characterizes the physical world as a limited projection of the forms also referred to as ideal categories. Later, Carl Jung introduced the concept of universal archetypes, described as a collective unconscious, in which he relates to Plato's forms by describing the forms as Jungian version of the Platonian archetypes \cite{10.2307/43797850}. Employing the theoretical concept of archetypes to political and ideological polarization, the archetypes could be interpreted as genuine ideologies, while the ideological advocates can be expressed as a mixture of distinct ideologies.

Ideological polarization is the distance between policy preferences, typically of elites taking extreme stands on issues whereas the electoral behavior is denoted affective polarization. When these extremes are portrayed as existential in the media, they typically form an "us-versus-them"-mindset \citep{dagnes2019us}. From a social network perspective, the process of polarization has been described to occur when "homophily and influence become self-reinforcing when the attraction to those who are similar and
differentiation from those who are dissimilar entail greater openness to influence. The result is network autocorrelation—the tendency for people to resemble their network neighbors" \citep{dellaposta2015liberals}.

To better capture ideological polarization, we turn to signed networks. Signed networks reflect complex social polarization better than unsigned networks because they capture positive, negative, and neutral relationships between entities. The study of signed networks goes back to the ’50s and was motivated by friendly and hostile social relationships \citep{harary}. Since then they have been used to study networks of Twitter users \citep{twitter} and US Congress members \citep{uscongress}, two examples of polarized social networks \citep{pol_twitter,pol_uscongress}.

In this paper, we focus on polarization as extreme positions and argue that the multi-polarity of "us-versus-them" reinforced by homophily and influence can be characterized by a latent position model (i.e., the latent distance model \citep{Hoff}) of networks confined to a constrained social space formed by a polytope, what we denote a sociotope. As such, the corners of the sociotope define distinct aspects (i.e., poles) formed by polarized networks' tendencies to self-reinforce homophily by positive ties driving those who are similar close as opposed to those that are negatively tied being repelled. This can be revealed in terms of the important multiple poles of social network defining corners of such sociotope. Within these corners, positive interactions between nodes place them in close proximity in space thereby accounting for homophily while negative interactions "push" nodes far apart (towards opposing poles) yielding the "us-versus-them" effect.

The conceptual idea of polytopes as formed by pure types can be traced back to Plato's forms, which characterize the physical world as a limited projection of the forms also referred to as ideal categories. Later, Carl Jung introduced the concept of universal archetypes, described as a collective unconscious, in which he related to Plato's forms by describing the forms as a Jungian version of the Platonian archetypes \citep{10.2307/43797850}. Employing the theoretical concept of archetypes to political and ideological polarization, the archetypes could be interpreted as genuine ideologies, while the ideological advocates can be expressed as a mixture of distinct ideologies.

Archetypal Analysis (AA) is a prominent framework for extracting polytopes in tabular data. AA was originally proposed by \cite{cutler1994a} as an unsupervised learning method that favors distinct aspects, archetypes, of the data in which observations are characterized by convex combinations (i.e., mixtures) of these archetypes as opposed to clustering procedures extracting prototypical observations \citep{5589222}. AA has previously been used to model societal conflicts in Europe \citep{https://doi.org/10.1111/jcms.13342}. However, given that AA was proposed for tabular data, the applications are currently restricted to non-relational data. Thus, whereas the characterization of data in terms of distinct aspects and polytopes has a long history, such representation learning approaches have not previously been considered in the context of network analysis for the extraction of polarization by several extremes. 

In the last years, representation learning of signed graphs has gathered substantial attention, with applications such as signed link prediction \citep{sign_link}, and community detection \citep{sign_com}. Initial works extended the prominent random walks framework \citep{deepwalk-perozzi14,node2vec-kdd16} to the analysis of signed networks. \textsc{SIDE} \citep{side} exploits truncated random walks on the signed graph with interaction signs for each node pair inferred based on balance theory \citep{balance_theory}. Balance theory is a socio-psychological theory admitting four rules: “The
friend of my friend is my friend," “The friend of my enemy is my
enemy," “The enemy of my friend is my enemy," and “The enemy of
my enemy is my friend." \textsc{POLE} \citep{pole}, also utilizes balance theory-based signed random walks to construct an auto-covariance similarity which is used to obtain the embedding space. Neural networks have also been adopted for the analysis of signed networks. Both \textsc{SiNE} \citep{sine} and \textsc{SIGNet} \citep{signet} combine balance theory and multi-layer neural networks to learn the network embeddings while SIGNet uses targeted node sampling to provide scalable inference. In addition, graph neural networks have also been studied in the context of signed graphs. More specifically, \textsc{SiGAT} \citep{sigat} and \textsc{SDGNN} \citep{SDGNN} combine balance and status theory with graph attention to learn signed network embeddings. The status theory is another important socio-psychological theory for directed relationships where for a source and a target node, a positive directed connection assumes a higher status of the target, i.e. $\{\text{status(target)}>\text{status(source)}\}$, while the inequality is opposite for a negative connection. Lastly, \textsc{SLF} \citep{slf} learns multiple latent factors of the signed network, modeling positive, negative, and neutral, as well as the absence of a relationship between node pairs.

A prominent approach for graph representation learning is the Latent Distance Model \citep{Hoff} in which the tendency of nodes to connect is defined in terms of their proximity in latent space. Notably, the LDM can express the properties transitivity (\textit{"a friend of a friend is a friend"}) and homophily (\textit{"akin nodes tend to have links"}). Recently, it has been shown that LDMs can account for the structure of networks in ultra-low dimensions \citep{nakis2022a,hmldm,pivem}. It has further been demonstrated that an LDM of one dimension can be used to extract bipolar network properties \citep{barbera2015tweeting}. 

For the modeling of signed networks for the characterization of polarization, we first present the Signed Latent Distance Model (\textsc{SLDM}). The model utilizes a likelihood function for weighted signed links based on the Skellam distribution \citep{jg1946frequency}. The Skellam distribution is the discrete probability distribution of the difference between two independent Poisson random variables. It was introduced by John Gordon Skellam to model the dynamics of populations \citep{jg1946frequency}. Since then it was used in medicine to model treatment measurements \citep{skellam_medecine}, sports results \citep{skellam_sport}, as well as, econometric studies \citep{skellam_econ}. Furthermore, we introduce the Signed relational Latent dIstance Model (\textsc{SLIM}) being able to characterize the latent social space in terms of extreme positions forming polytopes inspired by archetypal analysis enabling archetypal analysis for relational data, i.e. relational AA (RAA). We apply \textsc{SLDM} and \textsc{SLIM} %verify the models ability to uncover extremes %in synthetically generated networks reflecting polarization and 
on four real signed networks believed to reflect polarization and demonstrate how \textsc{SLIM} uncovers prominent distinct positions (poles). 
To the best of our knowledge, this is the first work to model signed weighted networks using the Skellam distribution and 
the first time AA has been extended to relational data by leveraging latent position modeling approaches for the characterization of polytopes in social networks. \noindent\textbf{The implementation is available at:} \href{https://github.com/Nicknakis/SLIM_RAA}{\textit{github.com/Nicknakis/SLIM\_RAA}}.

%% file: 3-proposed_model.tex
\section{PROPOSED METHODOLOGY}\label{sec:model}
% \subsection{Notation}
% %Notation reference https://arxiv.org/pdf/1901.06845.pdf 
% In the context of this paper, the notation is adopted from \cite{https://doi.org/10.48550/arxiv.1901.06845}. Let $\mathcal{G}=(\mathcal{V},\mathcal{E}, \mathcal{\sigma})$ be an undirected signed graph, $\mathcal{V}$ corresponds to the node set, $\mathcal{E} \subseteq \mathcal{V} \times \mathcal{V}$ shows the edge set and $\sigma$ is the sign function $\sigma : \mathcal{E}\rightarrow \{-1, +1\}$. Furthermore, $\mathbf{Y}_{N\times N}=(y_{i,j})\in \{-1,0,1\}^{N\times N}$ denotes the signed adjacency matrix of the graph, with each entry $y_{i,j}$ defined by:
% \begin{equation}
%   y_{i,j}=\begin{cases}
%     \sigma_{i,j}, & \text{if $(i,j)\in\mathcal{E}$}.\\
%     0, & \text{otherwise}.
%   \end{cases}
%   \label{signedAdjacencyMatrix}
% \end{equation}
Let $\mathcal{G}=(\mathcal{V}, \mathcal{Y})$ be a \textit{signed graph} where $\mathcal{V}=\{1,\ldots,N\}$ denotes the set of vertices and $\mathcal{Y}:\mathcal{V}^2 \rightarrow \mathsf{X}\subseteq\mathbb{R}$ is a map indicating the weight of node pairs, such that there is an edge $(i,j)\in\mathcal{V}^2$ if the weight $\mathcal{Y}(i,j)$ is different from $0$. In other words, $\mathcal{E}:=\{(i,j)\in\mathcal{V}^2:\mathcal{Y}(i,j)\not=0\}$ indicates the set of edges of the network. Since many real networks consist of only integer-valued edges, in this paper, we set $\mathsf{X}$ to $\mathbb{Z}$, and we will call the graph \textit{undirected} if the pairs $(i,j)$ and $(j,i)$ represent the same link. (The directed case is provided in the supplementary materials.) For simplicity, $y_{ij}$ denotes each edge weight.

\subsection{The Skellam Latent Distance Model (SLDM)}
Our main purpose is to learn latent node representations $\{\mathbf{z}_i\}_{i\in\mathcal{V}}\in\mathbb{R}^{K}$ in a low dimensional space for a given signed network $\mathcal{G}=(\mathcal{V}, \mathcal{Y})$ ($K \ll |\mathcal{V}|$). Therefore, the edge weights can take any integer value to represent the positive or negative tendencies between the corresponding nodes. We model these signed interactions among the nodes using  
 the Skellam distribution \citep{jg1946frequency}, which can be formulated as the difference of two independent Poisson-distributed random variables ($y=N_1 - N_2\in\mathbb{Z}$) with respect to the rates $\lambda^{+}$ and $\lambda^{-}$: 
\begin{align*}
P(y|\lambda^{+},\lambda^{-}) = e^{-(\lambda^{+}+\lambda^{-})}\left(\frac{\lambda^{+}}{\lambda^{-}}\right)^{y/2}\mathcal{I}_{|y|}\left(2\sqrt{\lambda^{+}\lambda^{-}}\right),
\end{align*}
where $N_1 \sim Pois(\lambda^{+})$ and $N_2 \sim Pois(\lambda^{-})$, and $\mathcal{I}_{|y|}$ is the modified Bessel function of the first kind and order $|y|$. To the best of our knowledge, the Skellam distribution has not been adapted before for modeling the network likelihood. More specifically, we propose a novel latent space model utilizing the Skellam distribution by adopting the latent distance model, which was proposed originally for undirected, and unsigned binary networks as a logistic regression model \citep{Hoff}. It was later extended to multiple generalized linear models \citep{bilinear}, including the Poisson regression model for integer-weighted networks. We can formulate the negative log-likelihood of a latent distance model under the Skellam distribution as:
\begin{align*}
\mathcal{L}(\mathcal{Y}) &:=\log p(y_{ij}|\lambda^{+}_{ij},\lambda^{-}_{ij}) \\
&= \sum_{i<j}{(\lambda^{+}_{ij}+\lambda^{-}_{ij})} - \frac{y_{ij}}{2}\log\left(\frac{\lambda^{+}_{ij}}{\lambda^{-}_{ij}}\right)-\log(I_{ij}^{*}),
\end{align*}
where $I_{ij}^{*} := \mathcal{I}_{|y_{ij}|}\left(2\sqrt{\lambda^{+}_{ij}\lambda^{-}_{ij}}\right)$. As it can be noticed, the Skellam distribution has two rate parameters, and we consider them to learn latent node representations $\{\mathbf{z}_i\}_{i\in\mathcal{V}}$ by defining them as follows:
\begin{align}
\lambda_{ij}^{+} &= \exp\big(\gamma_{i} + \gamma_{j} - ||\mathbf{z}_i-\mathbf{z}_j||_2\big)\label{eq:rate1},
\\
\lambda_{ij}^{-} &= \exp\big(\delta_{i} + \delta_{j} + ||\mathbf{z}_i-\mathbf{z}_j||_2\big),
\label{eq:rate2}
\end{align}
where the set $\{\gamma_i,\delta_i\}_{i\in\mathcal{V}}$ denote the node-specific random effect terms, and $||\cdot||_2$ is the Euclidean distance function. More specifically, $\gamma_i,\gamma_j$ represent the "social" effects/reach of a node and the tendency to form (as a receiver and as a sender, respectively) positive interactions, expressing positive degree heterogeneity (indicated by $+$ as a superscript of $\lambda$). In contrast, $\delta_i,\delta_j$ provide the "anti-social" effect/reach of a node to form negative connections, and thus models negative degree heterogeneity (indicated by $-$ as a superscript of $\lambda$). 

By imposing standard normally distributed priors elementwise on all model parameters $\bm{\theta}=\{\boldsymbol{\gamma},\bm{\delta}, \mathbf{Z}\}$, i.e., $\theta_i\sim \mathcal{N}(0,1)$, We define a maximum a posteriori (MAP) estimation over the model parameters, via the loss function to be minimized (ignoring constant terms):
\begin{align}\label{eq:loss_sk}
\begin{split}
    Loss =& \sum_{i<j}\Bigg( \lambda_{ij}^{+}+\lambda_{ij}^{-} -\frac{y_{ij}}{2}\log\left( \frac{\lambda_{ij}^{+} }{\lambda_{ij}^{-}}\right)\Bigg) 
    \\ 
    -& \sum_{i<j}\log I_{|y_{ij}|}\Big (2\sqrt{\lambda_{ij}^{+}\lambda_{ij}^{-}}\Big )
    \\
    +& \frac{\rho}{2}\Big(||\mathbf{Z}||_F^2+||\bm{\gamma}||_F^2+||\bm{\delta}||_F^2\Big),
    \end{split}
\end{align}
where $||\cdot||_F$ denotes the Frobenius norm. In addition, $\rho$ is the regularization strength with $\rho=1$ yielding the adopted normal prior with zero mean and unit variance. Importantly, by setting $\lambda_{ij}^{+}$ and $\lambda_{ij}^{-}$ based on Eq. \eqref{eq:rate1}
and \eqref{eq:rate2}, the model effectively makes positive (weighted) links attract and negative (weighted links) deter nodes from being in proximity of each other. %We leave more complex model formulations as a discussions.
\subsection{Archetypal Analysis}
Archetypal Analysis (AA) \citep{cutler1994a,5589222} is an approach developed for the modeling of observational data in which the data is expressed in terms of convex combinations of characteristics (i.e. archetypes). The definition of the embedded data points is given as follows:
\begin{eqnarray}
    \mathbf{X}\approx \mathbf{XCZ}\quad
    \text{s.t. }\boldsymbol{c}_d\in \Delta^{N} \text{ and } \mathbf{z}_j\in \Delta^{K}
\end{eqnarray}
where $\Delta^{P}$ denotes the standard simplex in $(P+1)$ dimensions such that $\mathbf{q}\in\Delta^{P}$ requires  $q_i\geq 0$ and $\|\mathbf{q}\|_1=1$ (i.e. $\sum_iq_i=1$). Notably, the archetypes given by the columns of $\mathbf{A}=\mathbf{X}\mathbf{C}$ define the corners of the extracted polytope as convex combinations of the observations, whereas $\mathbf{Z}$ define how each observation is reconstructed as convex combinations of the extracted archetypes.

Whereas archetypal analysis constrains the representation to the convex hull of the data, other approaches to model pure/ideal forms have been Minimal Volume (MV) approaches defined by
\begin{eqnarray}
\mathbf{X}\approx \mathbf{AZ}\quad \text{s.t. } vol(\mathbf{A})= v \text{ and } \mathbf{z}_j\in \Delta^{K},
\end{eqnarray}
in which $vol(\mathbf{A})$ defines the volume of $\mathbf{A}$. When $\boldsymbol{A}$ is a square matrix this can be defined by $vol(\mathbf{A})=|det(\mathbf{A})|$, see
also \cite{hart2015inferring,zhuang2019regularization} for a review on such end-member extraction procedures. A strength is that, as opposed to AA, the approach does not require the presence of pure observations, however, a drawback is a need for regularization tuning to define an adequate volume \citep{zhuang2019regularization} whereas the exact computation of the volume of general polytopes requires the computation of determinants of the sum of all simplices defining the polytope \citep{bueler2000exact}.
%, i.e.
%\begin{equation}
%vol(\mathbf{A})=\sum_{i=1}^I \frac{|det(\mathbf{a}_{s_i+}-\mathbf{a}_{s_i(0)},\ldots, \mathbf{a}_{s_i(D)}-\mathbf{a}_{s_i(0)})|}{D!}
%\end{equation}
%in which $I$ denotes the total number of simplices and $s_i(0:D)$ indexes the corners used to define the vertices of the $i^{th}$ simplex. This computations becomes expensive in high-dimensions
Importantly, Archetypal Analysis and Minimal volume extraction procedures have been found to identify latent polytopes defining trade-offs in which vertices of the polytopes represent maximally enriched distinct aspects (archetypes), allowing identification of tasks or prominent roles the vertices of the polytope represent \citep{shoval2012evolutionary,hart2015inferring}. Due to the computational issues of regularizing high-dimensional volumes and the need for careful tuning of such regularization parameters, we presently focus on polytope extraction as defined through the AA formulation rather than the MV formulation.

\subsection{A Generative Model of Polarization}\label{generative}

Considering a latent space for the modeling of polarization, we presently extend the Skellam LDM and define polarization as extreme positions (pure forms/archetypes) that optimally represent the social dynamics observed in terms of the induced polytope - what we denote a sociotope, in which each observation is a convex combination of these extremes. In particular, we characterize polarization in terms of extreme positions in a latent space defined as a polytope akin to AA and MV. 

In our generative model of polarization,
we further suppose that the bias terms introduced in the definitions of the Poisson rates, $(\lambda_{ij}^{+},\lambda_{ij}^{-})$, are normally distributed. Since latent representations $\{\mathbf{z}_i\}_{i\in\mathcal{V}}$ according to AA and MV lie in the standard simplex set $\Delta^K$, we further assume that they follow a Dirichlet distribution. Formally, we can summarize the generative model as follows:
\begin{align*}
\gamma_i &\sim \mathcal{N}(\mu_\gamma,\sigma_\gamma^2) && \forall i\in\mathcal{V}, \\
\delta_i &\sim \mathcal{N}(\mu_\delta,\sigma_\delta^2) && \forall i\in\mathcal{V}, \\
\mathbf{a}_{k} &\sim \mathcal{N}(\boldsymbol{\mu}_A, \sigma_A^2\mathbf{I}) && \forall k\in\{1,\ldots,K\},\\
\mathbf{z}_i &\sim Dir(\bm{\alpha}) && \forall i\in\mathcal{V}, \\
\lambda_{ij}^{+} &= \exp\big(\gamma_{i} + \gamma_{j} - \|\mathbf{A}(\mathbf{z}_i-\mathbf{z}_j)\|_2\big),\\
\lambda_{ij}^{-} &= \exp\big(\delta_{i} + \delta_{j} + \|\mathbf{A}(\mathbf{z}_i-\mathbf{z}_j)\|_2\big),\\
y_{ij} &\sim Skellam(\lambda^{+}_{ij},\lambda_{ij}^{-}) && \forall (i,j)\in\mathcal{V}^2.
\end{align*}
According to the above generative process, positive ($\boldsymbol{\gamma}$) and negative ($\boldsymbol{\delta}$) random effects for the nodes are first drawn, upon which the location of extreme positions $\mathbf{A}$ (i.e., corners of the polytope denoted archetypes) are generated. In addition, as the dimensionality of the latent space increases linearly with the number of archetypes, i.e. $\boldsymbol{A}$ is a square matrix, with probability zero archetypes will be placed in the interior of the convex hull of the other archetypes. Subsequently, the node-specific convex combinations $\mathbf{Z}$ of the generated archetypes are drawn, and finally, the weighted signed link is generated according to the node-specific biases and distances between dyads within the polytope utilizing the Skellam distribution.

%\begin{figure*}[h]
%\centering
%\includegraphics[scale=0.3]{slim.pdf}
%\caption{Given an input a signed network for node $v$ the \textit{Signed Relational Distance Model} projects the node $v$ into the sociotope. We give an example, of three sociotopes with $K=2,3,8$ corners/pure forms.}
%\label{fig:train}
%\end{figure*}

\subsection{The Signed Relational Latent Distance Model}

For inference, we exploit how polytopes can be efficiently extracted using archetypal analysis. We, therefore, define the \textsc{S}igned \textsc{L}atent relational d\textsc{I}stance \textsc{M}odel (\textsc{SLIM}) by defining a relational archetypal analysis approach endowing the generative model a parameterization akin to archetypal analysis in order to efficiently extract polytopes from relational data defined by signed weighted networks. Specifically, we formulate the relational AA in the context of the family of LDMs, as:
\begin{align}
    \lambda_{ij}^{+} &=\exp \big( \gamma_{i} + \gamma_{j} - \|\mathbf{A} (\mathbf{z}_{i}-\mathbf{z}_{j})\|_{2}\big)
    \\ 
    &=\exp \big( \gamma_{i} + \gamma_{j} -\|\mathbf{RZC}(\mathbf{z}_{i}-\mathbf{z}_{j})\|_2\big).\label{LRPM_inensity_function_1}
    \\
\lambda_{ij}^{-} &=\exp \big( \delta_{i} + \delta_{j} + \|\mathbf{A} (\mathbf{z}_{i}-\mathbf{z}_{j})\|_{2}\big)
\\ 
&=\exp \big( \delta_{i} + \delta_{j} +\|\mathbf{RZC}(\mathbf{z}_{i}-\mathbf{z}_{j})\|_2\big).\label{LRPM_inensity_function_2}
\end{align}
%Such that the \textsc{LSRDM} is defined as
%\begin{eqnarray}
%\mathbf{Y}\approx \mathbf{\Lambda}^{+}-\mathbf{\Lambda}^{-},\boldsymbol{c}_d\in\Delta^{N},\text{ and } \mathbf{z}_i\in\Delta^K.
%\end{eqnarray}
Notably, in the AA formulation $\mathbf{X}=\mathbf{RZ}$ corresponds to observations formed by convex combinations $\mathbf{Z}$ of positions given by the columns of $\boldsymbol{R}^{K\times K}$. Furthermore, in order to ensure what is used to define archetypes $\mathbf{A}=\mathbf{XC}=\mathbf{RZC}$ corresponds to observations using these archetypes in their reconstruction $\mathbf{Z}$,
we define $\boldsymbol{C}\in \boldsymbol{R}^{N\times K}$ as a gated version of $\mathbf{Z}$ normalized to the simplex such that $\boldsymbol{c}_d\in\Delta^{N}$ by defining 
\begin{equation}
    c_{nd}=\frac{(\mathbf{Z}^\top\circ [\sigma(\mathbf{G})]^\top)_{nd}}{\sum_{n^\prime}(\mathbf{Z}^\top\circ [\sigma(\mathbf{G})]^\top)_{n^\prime d}}
\end{equation}
in which $\circ$ denotes the elementwise (Hadamard) product and $\sigma(\mathbf{G})$ defines the logistic sigmoid elementwise applied to the matrix $\boldsymbol{G}$. As a result, the extracted archetypes are ensured to correspond to the nodes assigned the archetype, whereas the location of the archetypes can be flexibly placed in space as defined by $\mathbf{R}$. By defining $\mathbf{z}_i=\operatorname{softmax}(\tilde{\mathbf{z}}_i)$ we further ensure $\mathbf{z}_i\in \Delta^K$.
% We further assume node embeddings $\{\mathbf{z}\}_{i\in\mathcal{V}}$ lie in the standard simplex set, $\Delta^K$.
%Thus, the product $\mathbf{ZCz}_{i}$ corresponds to the archetypal convex projection of the latent embedding of the ith node.-\textbf{Added by Christian (maybe redundant)} 

Importantly, the loss function of Eq. \eqref{eq:loss_sk} is adopted for the relational AA formulation forming the \textsc{SLIM}, with the prior regularization applied to the corners of the extracted polytope $\bm{A}=\mathbf{RZC}$ instead of the latent embeddings $\bm{Z}$ imposing a standard elementwise normal distribution as prior $a_{k,k^\prime}\sim \mathcal{N}(0,1)$. Furthermore, we impose a uniform Dirichlet prior on the columns of $\bm{Z}$, i.e. $(\bm{z}_i\sim Dir(\mathbf{1}_K)$, this only contributes constant terms to the joint distribution, and therefore the maximum a posteriori (MAP) optimization only constant terms. As a result, the loss function optimized is given by Eq. \eqref{eq:loss_sk} replacing $\|\mathbf{Z}\|_F^2$ with $\|\boldsymbol{A}\|_F^2$.

\textbf{Complexity analysis.} With \textsc{SLDM}/\textsc{SLIM} being distance models, they scale prohibitively as $\mathcal{O}(N^2)$ since the node pairwise distance matrix needs to be computed. This does not allow the analysis of large-scale networks. For that, we adopt an unbiased estimation of the log-likelihood through random sampling. More specifically, gradient steps are based on the log-likelihood of the block formed by a sampled (per iteration and with replacement) set $S$ of network nodes. This makes inference scalable defining an $\mathcal{O}(S^2)$ space and time complexity. More options for scalable inference of distance models have also been proposed in \cite{nakis2022a,case_control}.

%% file: 4-experiments.tex
\section{RESULTS AND DISCUSSION}\label{sec:experiments}
We extensively evaluate the performance of our proposed methods by comparing them to the prominent GRL approaches designed for signed networks. All experiments regarding \textsc{SLDM}/\textsc{SLIM} have been conducted on an $8$ GB NVIDIA RTX $2070$ Super GPU. In addition, we adopted the Adam optimizer \cite{kingma2017adam} with learning rate $\text{lr}=0.05$ and for $5000$ iterations. The sample size for the node set was chosen as approximately $3000$ nodes for all networks. The initialization of the \textsc{SLDM}/\textsc{SLIM} frameworks is deterministic and based on the spectral decomposition of the normalized Laplacian (more details are provided in the supplementary).

\begin{figure*}[!h]
\centering
\begin{subfigure}{.2\linewidth}
  \centering
  \includegraphics[width=\linewidth]{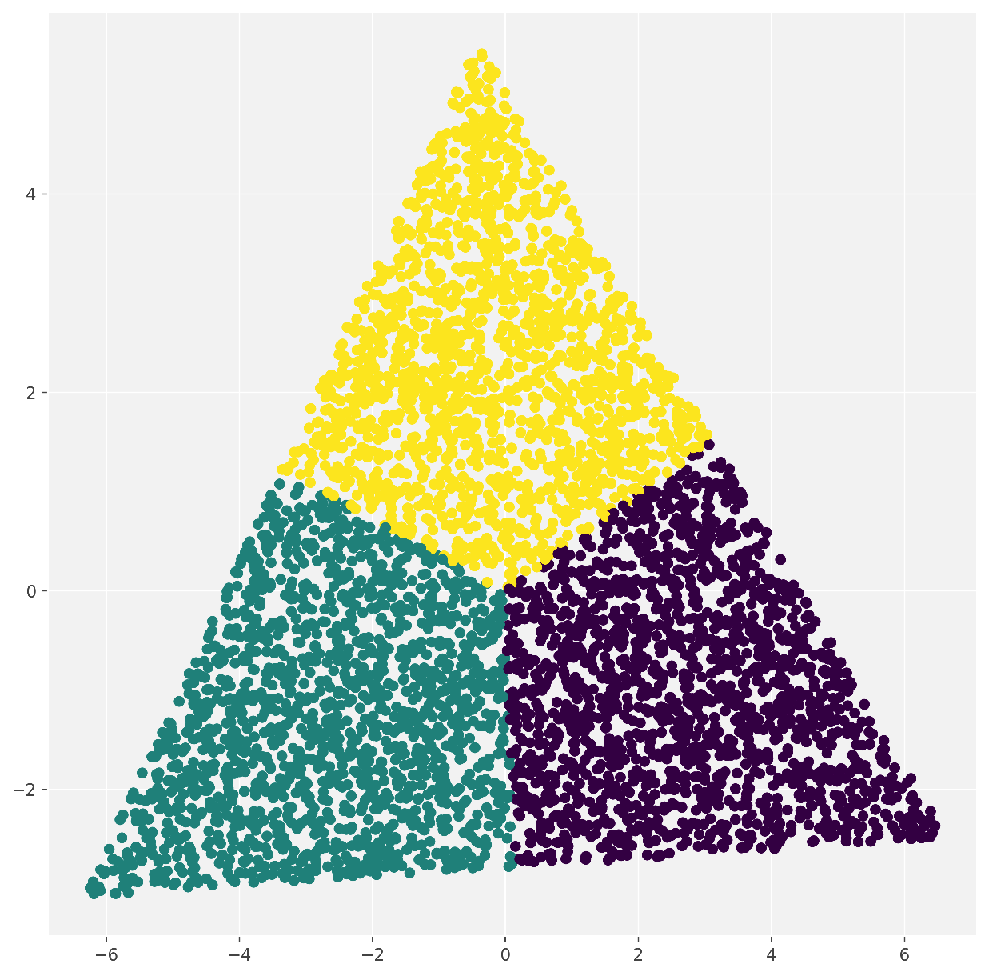}
  \caption{\scriptsize Ground Truth}
  \label{fig:sub1}
\end{subfigure}%
\hfill
\begin{subfigure}{.2\linewidth}
  \centering
  \includegraphics[width=\linewidth]{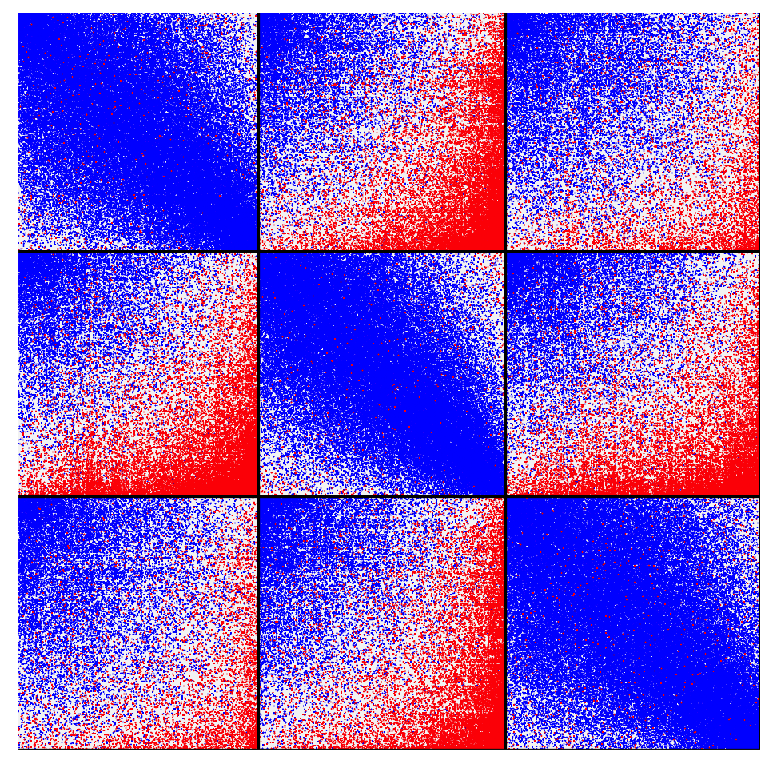}
  \caption {\scriptsize$(.017,77,23)$}
  \label{fig:sub2}
\end{subfigure}
\hfill
\begin{subfigure}{.2\linewidth}
  \centering
  \includegraphics[width=\linewidth]{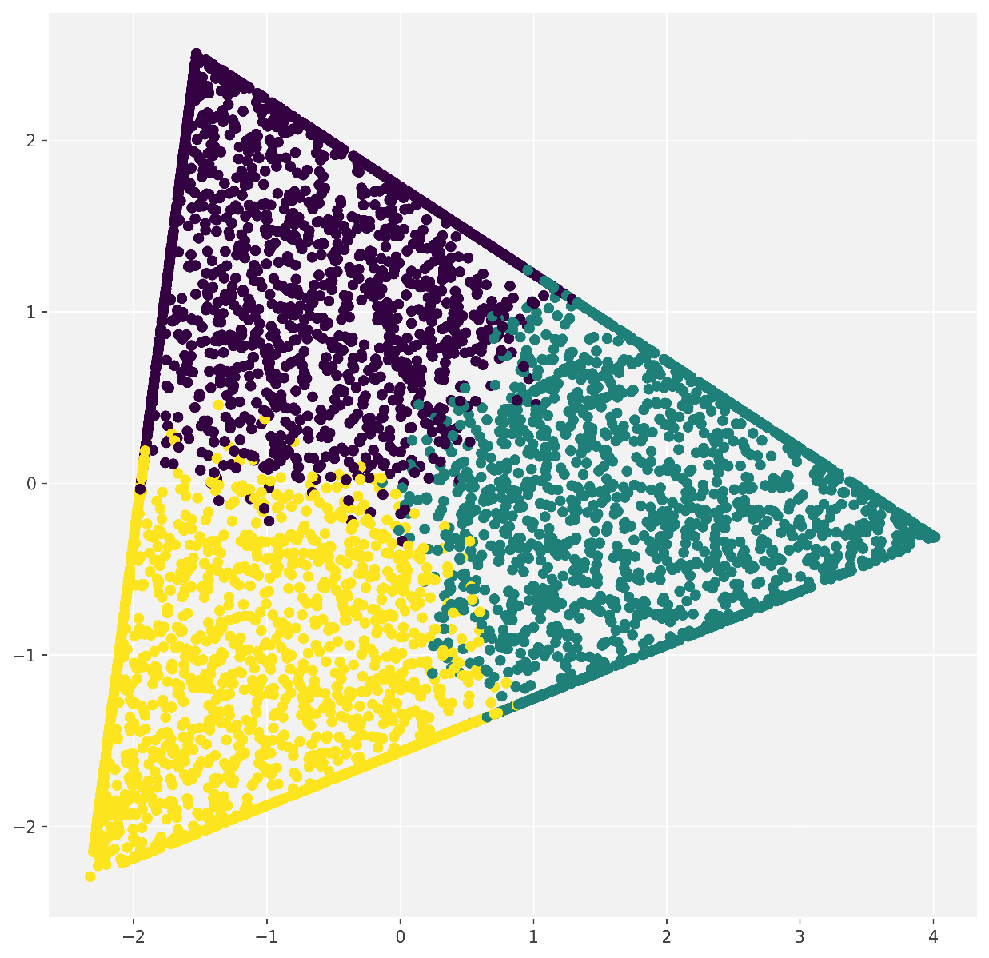}
  \caption{\scriptsize Inferred space}
  \label{fig:sub3}
\end{subfigure}
\hfill
\begin{subfigure}{.2\linewidth}
  \centering
  \includegraphics[width=\linewidth]{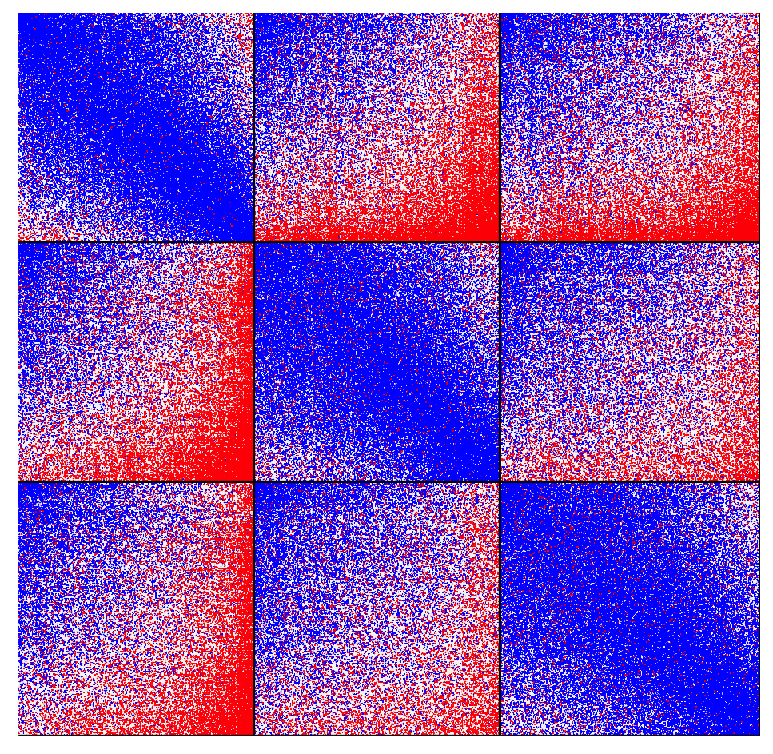}
  \caption{\scriptsize$(.018,73,27)$}
  \label{fig:sub4}
\end{subfigure}
\centering
\begin{subfigure}{.2\linewidth}
  \centering
  \includegraphics[width=\linewidth]{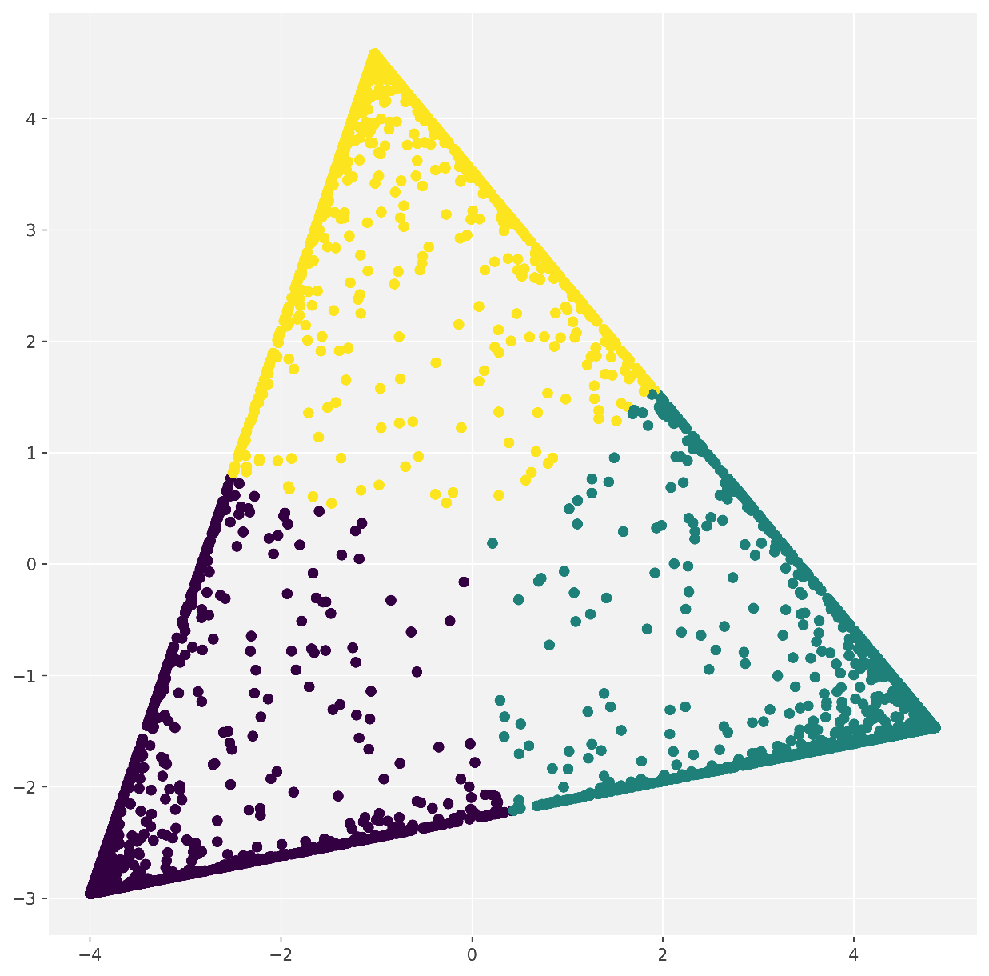}
  \caption{\scriptsize Ground Truth}
  \label{fig:sub5}
\end{subfigure}%
\hfill
\begin{subfigure}{.2\linewidth}
  \centering
  \includegraphics[width=\linewidth]{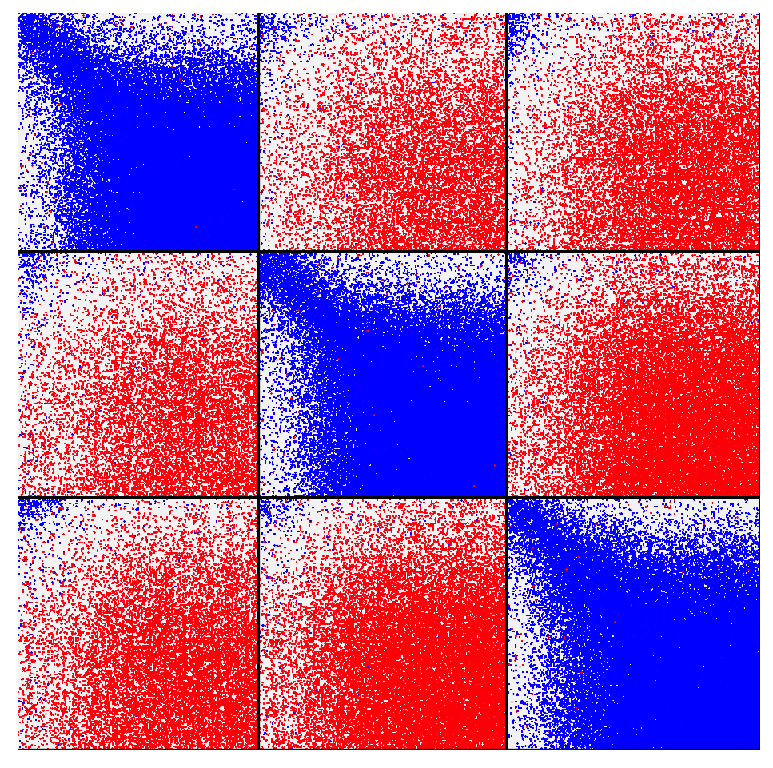}
  \caption {\scriptsize$(.012,63,37)$}
  \label{fig:sub6}
\end{subfigure}
\hfill
\begin{subfigure}{.2\linewidth}
  \centering
  \includegraphics[width=\linewidth]{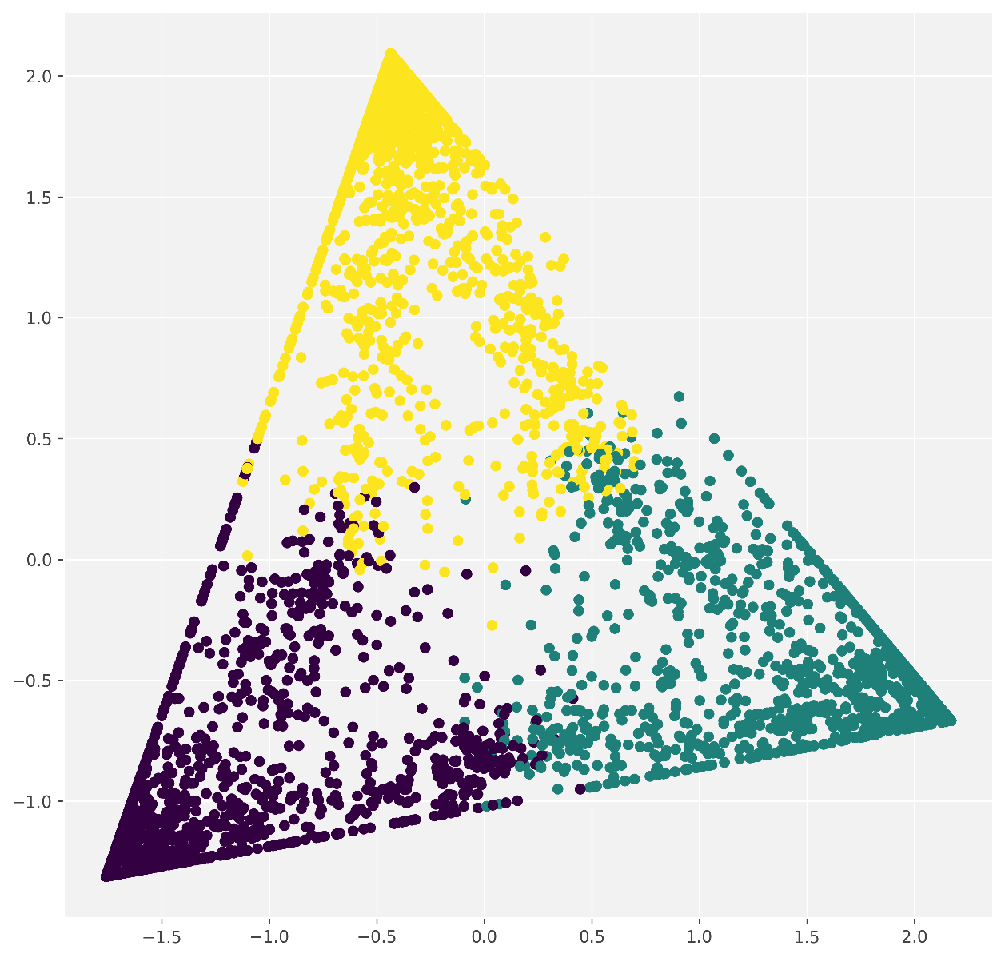}
  \caption{\scriptsize Inferred space}
  \label{fig:sub7}
\end{subfigure}
\hfill
\begin{subfigure}{.2\linewidth}
  \centering
  \includegraphics[width=\linewidth]{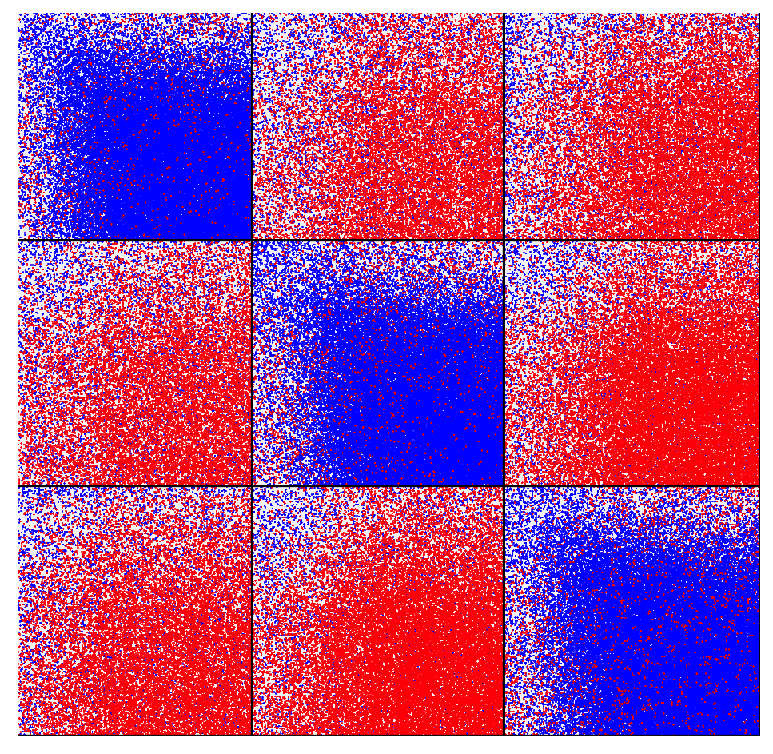}
  \caption{\scriptsize$(.014,59,41)$}
  \label{fig:sub8}
\end{subfigure}
\caption{Two artificially generated networks with different levels of polarization \{$\mathbf{z}_i \sim Dir(\bm{1})$ (top row), and $\mathbf{z}_i \sim Dir(0.1\cdot\bm{1})$ (bottom row)\}. Both size $N=5000$ nodes and $K=3$ archetypes. The first column shows the first two principal components of the original latent space $\tilde{\bm{Z}}=\bm{A}\mathbf{Z}$, the second column the original adjacency matrix, while the parenthesis shows the network statistics as: (density,\% of positive (blue) links,\% of negative (red) links). The third column displays the first two principal components of the inferred latent space, and the fourth column is the \textsc{SLIM} generated network based on inferred parameters. All network adjacency matrices are ordered based on $\mathbf{z}_i$, in terms of maximum archetype membership and internally according to the magnitude of the corresponding archetype most used for their reconstruction.}
\label{fig:art}
\end{figure*}

\textbf{Artificial networks.} We first, introduce experiments on artificial networks, as generated by the generative process described in Section \ref{generative}. We create two networks expressing different levels of polarization. Results are presented in Fig. \ref{fig:art}. More specifically, sub-Figs \ref{fig:sub1} and \ref{fig:sub5} show the ground truth latent spaces generating the networks with adjacency matrices as shown by sub-Figs \ref{fig:sub2} and \ref{fig:sub6}, respectively. The inferred latent spaces of the two networks are provided in sub-Figs \ref{fig:sub3} and \ref{fig:sub7} where it is clear that the model successfully distinguishes the difference in the level of polarization of the two networks. We also verify the generated networks based on the inferred parameters given by sub-Figs \ref{fig:sub4} and \ref{fig:sub8}. We observe that the model successfully generates sparse networks accounting for the positive and negative link imbalance.

\textbf{Real networks.} We employed four networks of varying sizes and structures. (\textbf{i}) \textsl{Reddit} is constructed based on hyperlinks representing the directed connections between two communities in a social platform \citep{dataset_reddit}. (\textbf{ii}) \textsl{wikiRfA} and (\textbf{iii}) \textsl{wikiElec} are the election networks covering the different time intervals in which nodes indicate the users and the directed links show supporting, neutral, and opposing votes to be selected as an administrator on the Wikipedia platform \citep{dataset_wikirfa,dataset_wikielec}. Finally, (\textbf{iv}) \textsl{Twitter} is an undirected social network built on the corpus of tweets concerning the highly polarized debate about the reform of the Italian Constitution \citep{dataset_twitter}. 

In our experiments, we consider the greatest connected component of the networks, and if the original network is temporal, we construct the static network by summing the weights of the links through time. For the experiments performed on undirected graphs, we similarly combine directed links to obtain the undirected version of the networks.

\begin{table} %[htbp]
\centering
\caption{Network statistics; $|\mathcal{V}|$: \# Nodes, $ |\mathcal{Y}^+|$: \# Positive links, $|\mathcal{Y}^-|$: \# Negative links.}
\label{tab:my-table}
\resizebox{\columnwidth}{!}{%
\begin{tabular}{rcccc}
\toprule
 & $|\mathcal{V}|$ & $|\mathcal{Y}^{+}|$ & $|\mathcal{Y}^{-}|$ & Density \\\midrule
\textsl{Reddit} & 35,776 & 128,182 & 9,639 & 0.0001 \\
\textsl{Twitter} & 10,885 & 238,612 & 12,794 & 0.0021 \\
\textsl{wiki-Elec} & 7,117 & 81,277 & 21,909 & 0.0020 \\
\textsl{wiki-RfA} & 11,332 & 117,982 & 66,839 & 0.0014\\\bottomrule
\end{tabular}%
}
\end{table}

\begin{table*}[!t]
\centering
\caption{Area Under Curve (AUC-ROC) scores for representation size of $K=8$.}
\label{tab:auc_roc}
\resizebox{0.85\textwidth}{!}{%
\begin{tabular}{rccccccccccccccccccccccccc}\toprule
\multicolumn{1}{l}{} & \multicolumn{3}{c}{\textsl{WikiElec}} & \multicolumn{3}{c}{\textsl{WikiRfa}} & \multicolumn{3}{c}{\textsl{Twitter}}& \multicolumn{3}{c}{\textsl{Reddit}} \\\cmidrule(rl){2-4}\cmidrule(rl){5-7}\cmidrule(rl){8-10}\cmidrule(rl){11-13}
\multicolumn{1}{r}{Task} & $p@n$ & $p@z$ & $n@z$ & $p@n$ & $p@z$ & $n@z$ & $p@n$ & $p@z$ & $n@z$ & $p@n$ & $p@z$ & $n@z$  \\\cmidrule(rl){1-1}\cmidrule(rl){2-2}\cmidrule(rl){3-3}\cmidrule(rl){4-4}\cmidrule(rl){5-5}\cmidrule(rl){6-6}\cmidrule(rl){7-7}\cmidrule(rl){8-8}\cmidrule(rl){9-9}\cmidrule(rl){10-10}\cmidrule(rl){11-11}\cmidrule(rl){12-12}\cmidrule(rl){13-13}
\textsc{POLE}     &.809 &.896 &.853 &.904 &.921 &.767 &.965 &.902 &.922 &x &x &x\\
\textsc{SLF}    &\textbf{.888} &.954 &\textbf{.952} &\textbf{.971} &.963 &\underline{.961} &.914 &.877 &.968 &\textbf{.729} &\textbf{.955} &.968\\
\textsc{SiGAT}     &.874 &.775 &.754 &.944 &.766 &.792 &\textbf{.998} &.875 &.963 &\underline{.707} &.682 &.712\\
\textsc{SIDE}      &.728 &.866 &.895 &.869 &.861 &.908 &.799 &.843 &.910 &.653 &.830 &.892\\
\textsc{SigNet}    &.841 &.774 &.635 &.920 &.736 &.717 &.968 &.719 &.891 &.646 &.547 &.623\\
\midrule
\textsc{SLIM (ours)}  &.862 &\underline{.965} &.935 &.956 &\underline{.980} &.960 &\underline{.988} &\textbf{.963} &\textbf{.972} &.667 &\textbf{.955} &\textbf{.978}\\
\textsc{SLDM (ours)}     & \underline{.876} &\textbf{.969} &\underline{.936} &\underline{.963} &\textbf{.982} &\textbf{.963} &.986 &\underline{.962} &\textbf{.973} &.648 &\underline{.951} &\underline{.975}
\\\bottomrule    
\end{tabular}%
}
\end{table*}

\begin{table*}[!t]
\centering
\caption{Area Under Curve (AUC-PR) scores for representation size of $K=8$.}
\label{tab:auc_pr}
\resizebox{0.85\textwidth}{!}{%
\begin{tabular}{rccccccccccccccccccccccccc}\toprule
\multicolumn{1}{l}{} & \multicolumn{3}{c}{\textsl{WikiElec}} & \multicolumn{3}{c}{\textsl{WikiRfa}} & \multicolumn{3}{c}{\textsl{Twitter}}& \multicolumn{3}{c}{\textsl{Reddit}} \\\cmidrule(rl){2-4}\cmidrule(rl){5-7}\cmidrule(rl){8-10}\cmidrule(rl){11-13}
\multicolumn{1}{r}{Task} & $p@n$ & $p@z$ & $n@z$ & $p@n$ & $p@z$ & $n@z$ & $p@n$ & $p@z$ & $n@z$ & $p@n$ & $p@z$ & $n@z$  \\\cmidrule(rl){1-1}\cmidrule(rl){2-2}\cmidrule(rl){3-3}\cmidrule(rl){4-4}\cmidrule(rl){5-5}\cmidrule(rl){6-6}\cmidrule(rl){7-7}\cmidrule(rl){8-8}\cmidrule(rl){9-9}\cmidrule(rl){10-10}\cmidrule(rl){11-11}\cmidrule(rl){12-12}\cmidrule(rl){13-13}
\textsc{POLE}    &.929 &.922 &.544 &.927 &.937 &.779 &\underline{.998} &.932 &.668 &x &x &x\\
\textsc{SLF} &\textbf{.964} &.926 &\textbf{.787} &\textbf{.983} &.922 &.881 &.994 &.870 &.740 &\textbf{.966} &\underline{.956} &\textbf{.850}\\
\textsc{SiGAT}  &\underline{.960} &.724 &.439 &.969 &.646 &.497 &\textbf{.999} &.861 &.582 &\underline{.965} &.692 &.232\\
\textsc{SIDE}    &.907 &.779 &.608 &.920 &.806 &.739 &.974 &.831 &.469 &.957 &.820 &.614\\
\textsc{SigNet} &.944 &.670 &.298 &.950 &.572 &.417 &\underline{.998} &.647 &.248 &.956 &.510 &.083\\
\midrule
\textsc{SLIM (ours)}    &.953 &\underline{.956} &\underline{.785} &.973 &\underline{.969} &\underline{.907} &\textbf{.999} &\underline{.962} &\textbf{.813} &.958 &\textbf{.960} &\textbf{.850}\\
\textsc{SLDM (ours)}     &\underline{.960} &\textbf{.963} &\textbf{.787} &\underline{.977} &\textbf{.971} &\textbf{.912} &\textbf{.999} &\textbf{.963} &\underline{.809} &.954 &.955 &\underline{.846}
\\\bottomrule    
\end{tabular}%
}
\end{table*}

\textbf{Baselines.} We benchmark the performance of our proposed frameworks against five prominent graph representation learning methods, designed for the analysis of signed networks: (\textbf{i}) \textsc{POLE} \citep{pole} which learns the network embeddings by decomposing the signed random walks auto-covariance similarity matrix, (\textbf{ii}) \textsc{SLF} \citep{slf} learns embeddings that are the concatenation of two latent factors targeting positive and negative relations, (\textbf{iii}) \textsc{SiGAT} \citep{sigat} is a graph neural network approach using graph attention to learn node embeddings, (\textbf{iv}) \textsc{SIDE} \citep{side} is another random walk based method for signed networks, (\textbf{v}) \textsc{SigNet} \citep{signet} is a multi-layer neural network approach constructing a Hadamard product similarity to accommodate for signed proximity on the network pairwise relations.

\subsection{Link prediction}
We evaluate performance considering the link prediction task considering the ability of our model to predict links of disconnected network pairs which should be connected, as well as, infer the sign of these links (positive or negative). For this, we remove/hide $20\%$ of the total network links while preserving connectivity on the residual network. For the testing set, the removed edges are paired with a sample of the same number of node pairs that are not the edges of the original network to create zero instances. To learn the node embeddings, we make use of the residual network.

\textbf{Predictions and evaluation metrics.} For our methods we fit a logistic regression classifier on the concatenation of the corresponding Skellam rates and log-rates, as $\chi_{ij}=\big [\lambda_{ij}^{+},\lambda_{ij}^{-},\log \lambda_{ij}^{+},\log \lambda_{ij}^{-}\big ]$. Since our Skellam likelihood formulation relies both on the ratio and products of the rates, a concatenation can take advantage of a linear function of the rates, as well as, their ratio or product as allowed from the log transformation. For the baselines, we use five binary operators \{average, weighted L1, weighted L2, concatenate, Hadamard product\} to construct feature vectors. For each of these feature vectors, we fit a logistic regression model (except for the Hadamard product which is used directly for predictions). Since different operators provide different performances, for the baselines we choose the operator that returns the maximum performance per individual task. As a consequence of the class imbalances and the sparsity present in signed networks, we adopt robust evaluation metrics, such as area-under-curve of the receiver operating characteristic (AUC-ROC) and precision-recall (AUC-PR) curves. Lastly, we denote with "x" the performance of a baseline if it was unable to run due to high memory/runtime complexity.

\textbf{Link sign prediction.} In this setting, we utilize the link test set containing the negative/positive cases of removed connections. We then ask the models to predict the sign of the removed links. We denote the task of the link sign prediction task as $p@n$. In Table \ref{tab:auc_roc} we provide the AUC-ROC scores while in Table \ref{tab:auc_pr} the AUC-PR scores for the undirected case. Here we observe that our proposed models outperform the baselines in most networks while being competitive in the \textsl{Reddit} network against \textsc{SLF}. This specific baseline is the most competitive across networks showing high and consistent performance similar to \textsc{SLIM} and \textsc{SLDM}. Comparing now \textsc{SLIM} with \textsc{SLDM} we get mostly on-par results, verifying that constraining the model to a polytope still provides enough expressive capability as the unconstrained model while allowing for accurate extraction of "extreme" positions.

\textbf{Signed link prediction.} The second and more challenging task is to predict removed links against disconnected pairs of the network, as well as, infer the sign of each link correctly. For that, the test set is split into two subsets positive/disconnected and negative/disconnected. We then evaluate the performance of each model on those subsets. The tasks of signed link prediction between positive and zero samples are denoted as $p@z$ while the negative against zero is $n@z$. We summarize our results by presenting AUC-ROC and AUC-PR scores in  Table \ref{tab:auc_roc} and Table \ref{tab:auc_pr} respectively. Once more our models outperform the baselines in most networks and for both versions of signed link prediction. The \textsc{SLF} baseline is again the most competitive baseline yielding on-par results for \textsl{Reddit}.

\textbf{Directed networks.} Directed network results are provided in the supplementary. Since \textsc{SLF} has higher modeling capacity it outperforms the simple model formulation of \textsc{SLDM} and \textsc{SLIM}. For that, we explore and discuss formulations allowing for more capacity in the \textsc{SLDM}/\textsc{SLIM} model for the directed case (see supplementary).

\begin{figure}[!t]
\begin{subfigure}{0.45\columnwidth}
    \includegraphics[width=\textwidth]{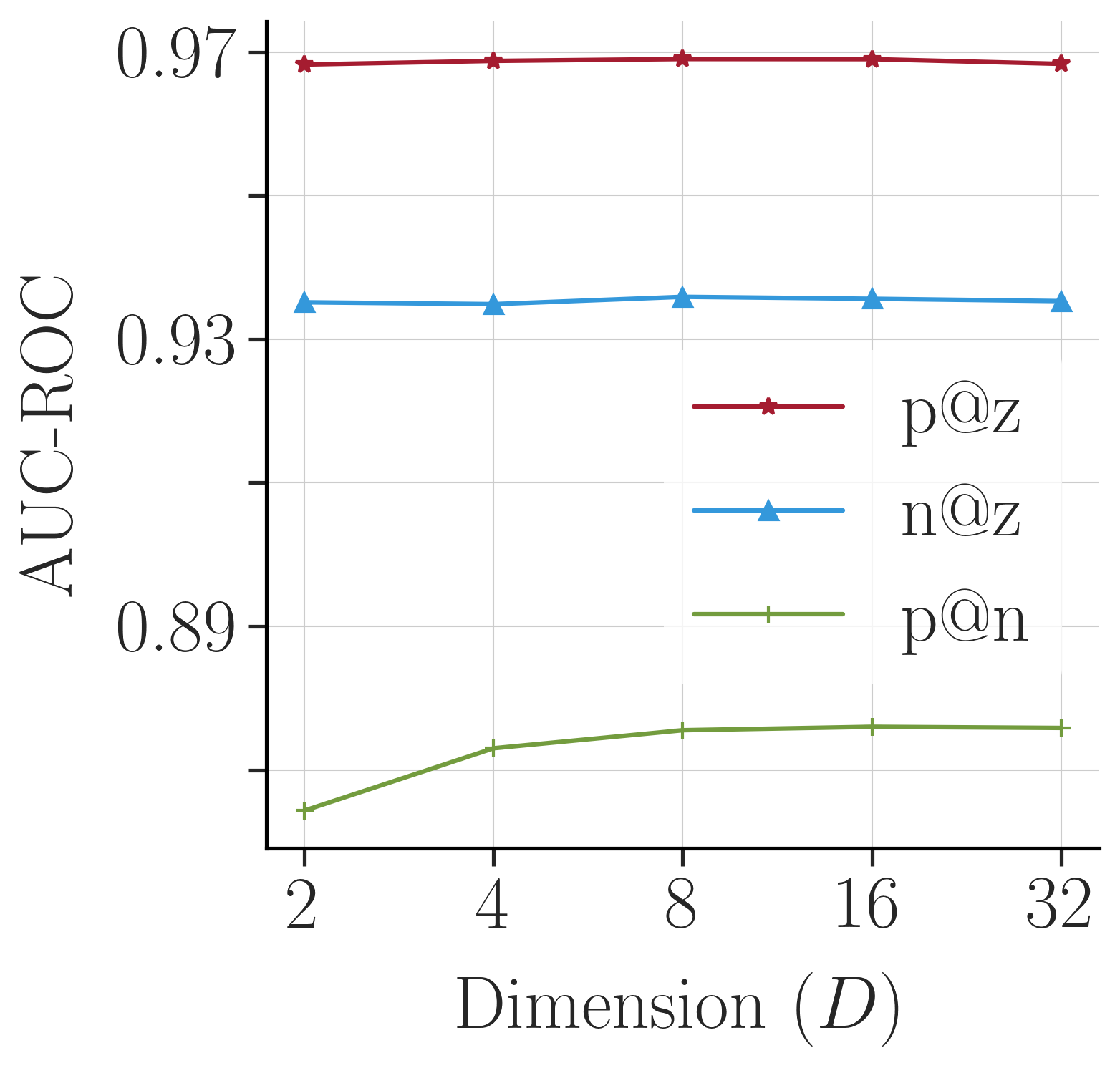}
    \caption{}
    \label{fig:roc_d}
\end{subfigure}
\hfill
\begin{subfigure}{0.45\columnwidth}
    \includegraphics[width=\textwidth]{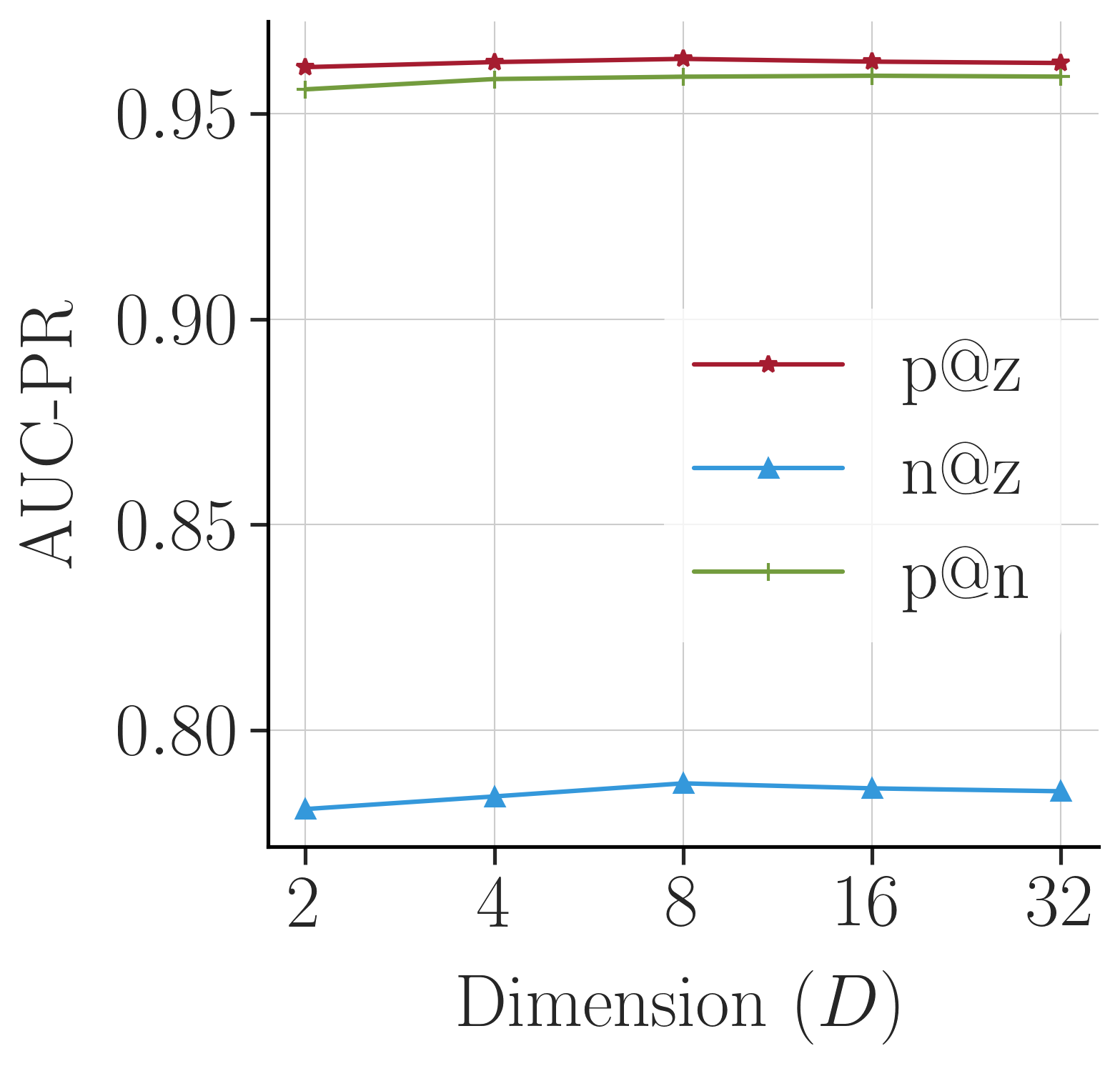}
        \caption{}
    \label{fig:pr_d}
\end{subfigure}
\caption{\textsl{wikiElec}: Performance of \textsc{SLIM} across dimensions for different tasks, (a) Area-Under-Curve Receiver Operating Characteristic scores, (b) Area-Under-Curve Precision-Recall scores. Both AUC-ROC and AUC-PR scores are almost constant across different dimensions }
\label{fig:roc_pr}
\end{figure}

\textbf{Effect of dimensionality.} In Figure \ref{fig:roc_pr}, we provide the performance across dimensions for the different downstream task and for the \textsl{wikiElec} dataset. We observe that both AUC-ROC and AUC-PR scores are almost constant across different dimensions (note that as $R^{K\times K}$ dimensions for the \textsc{SLIM} is given by the number of archetypes), showcasing that increasing the models' capacity (in terms of dimensions) does not have a significant effect on the performance of these downstream tasks (similar results were observed for all networks and most of the baselines).

\begin{figure*}
\centering
\begin{subfigure}{0.28\textwidth}
    \includegraphics[width=\textwidth]{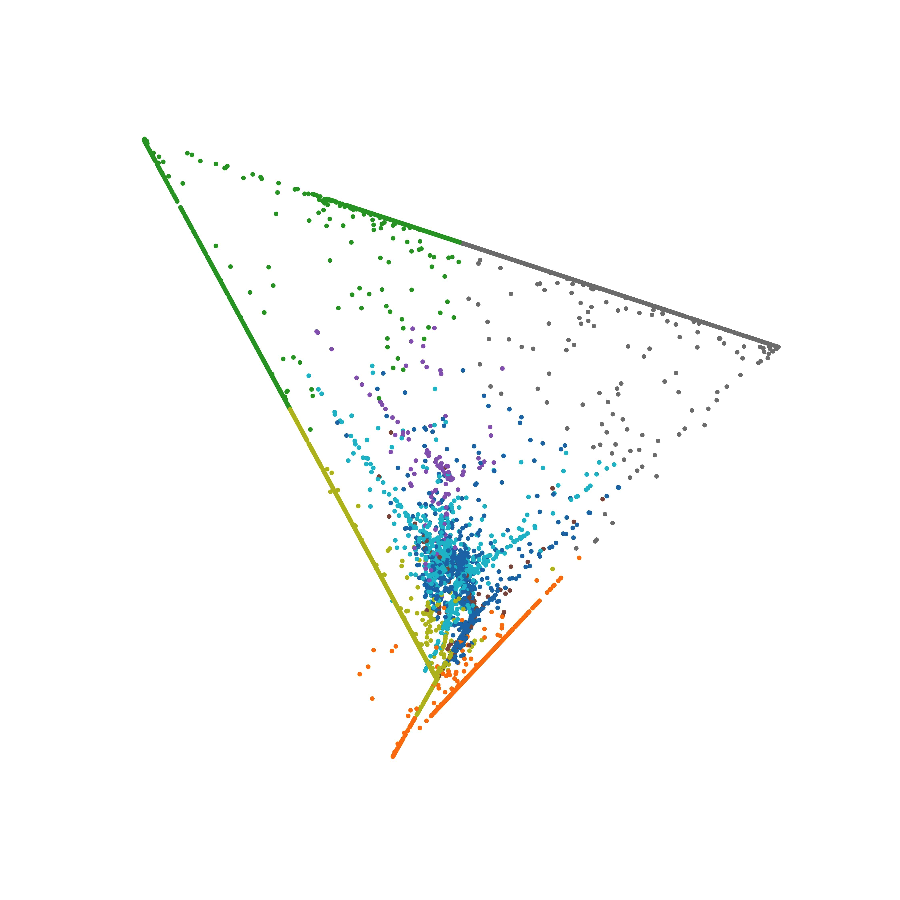}
    \caption{\textsl{WikiElec}}
    \label{fig:first}
\end{subfigure}
\hfill
\begin{subfigure}{0.28\textwidth}
    \includegraphics[width=\textwidth]{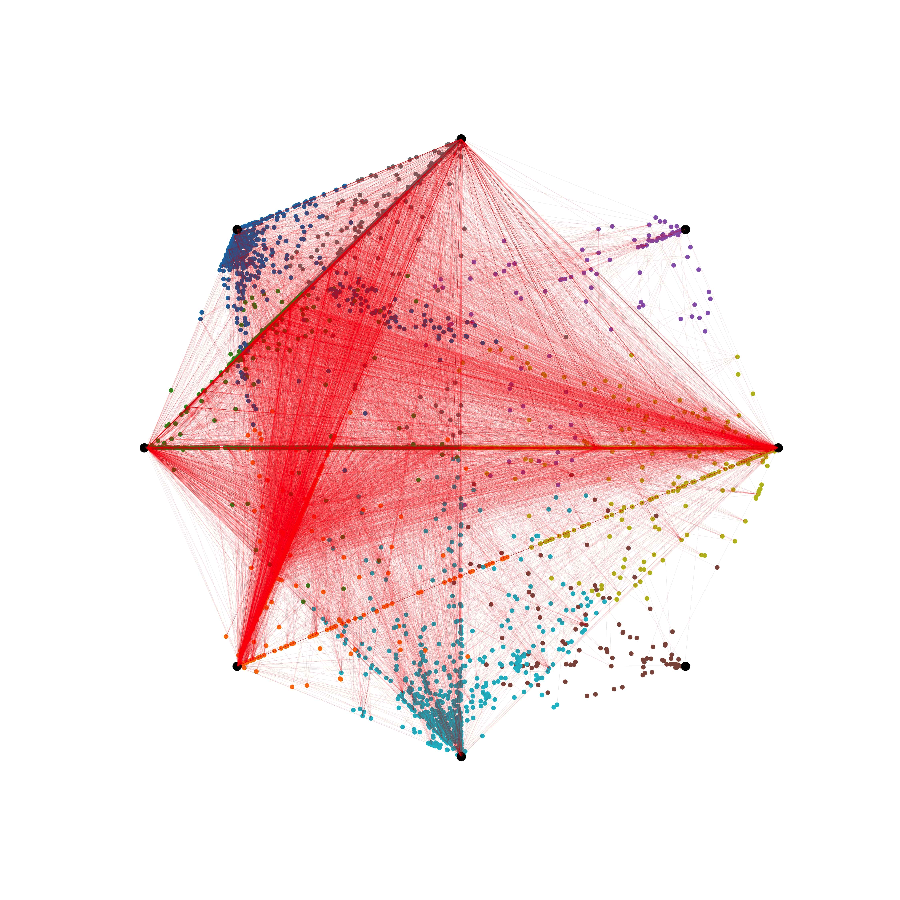}
    \caption{\textsl{WikiElec}}
    \label{fig:second}
\end{subfigure}
\hfill
\begin{subfigure}{0.28\textwidth}
    \includegraphics[width=\textwidth]{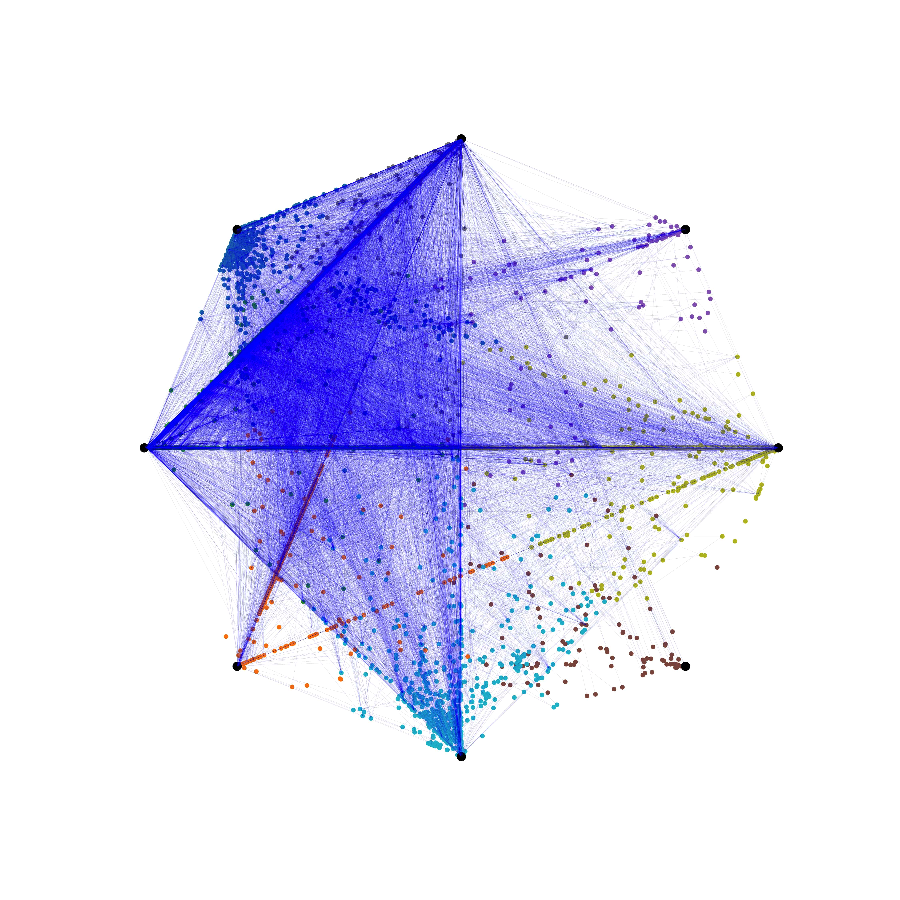}
    \caption{\textsl{WikiElec}}
    \label{fig:third}
\end{subfigure}
\begin{subfigure}{0.28\textwidth}
    \includegraphics[width=\textwidth]{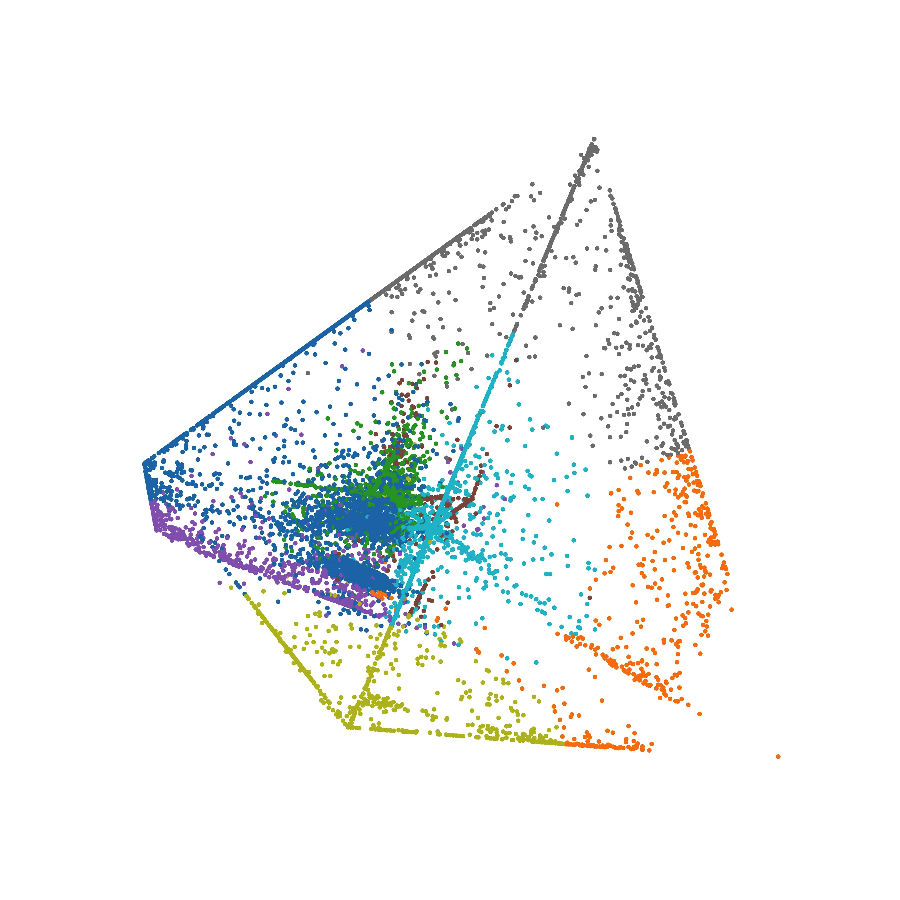}
    \caption{\textsl{WikiRfa}}
    \label{fig:first-2}
\end{subfigure}
\hfill
\begin{subfigure}{0.28\textwidth}
    \includegraphics[width=\textwidth]{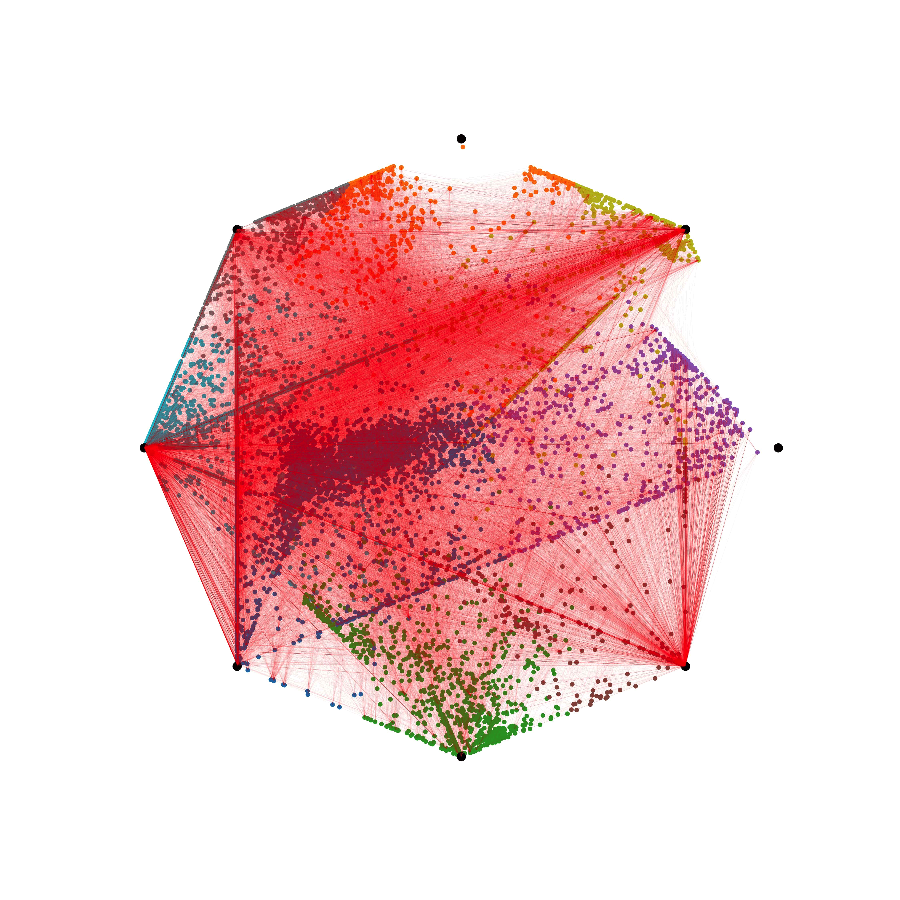}
    \caption{\textsl{WikiRfa}}
    \label{fig:second-2}
\end{subfigure}
\hfill
\begin{subfigure}{0.28\textwidth}
    \includegraphics[width=\textwidth]{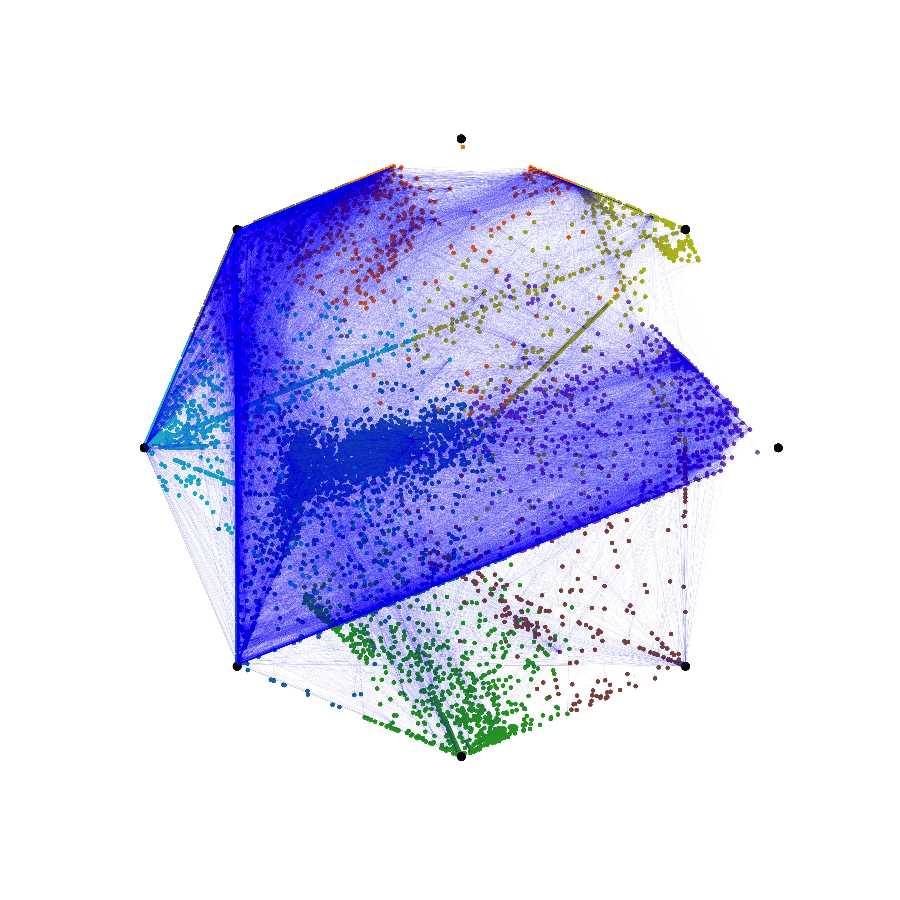}
    \caption{\textsl{WikiRfa}}
    \label{fig:third-2}
\end{subfigure}
\begin{subfigure}{0.28\textwidth}
    \includegraphics[width=\textwidth]{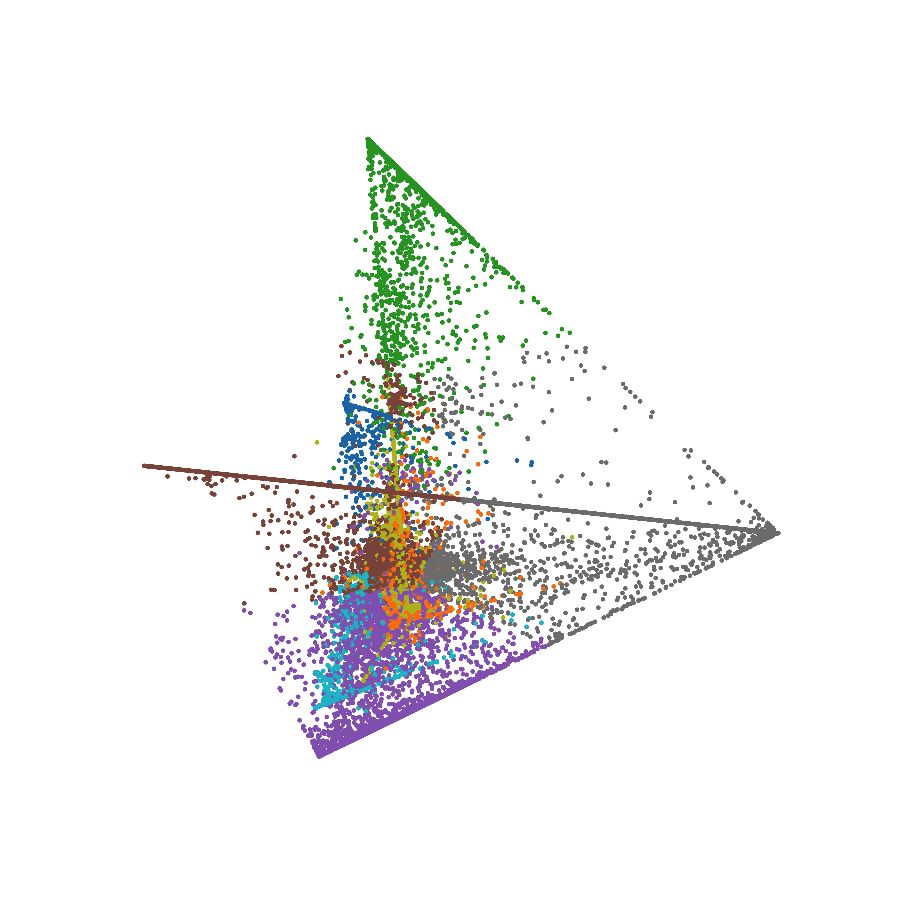}
    \caption{\textsl{Reddit}}
    \label{fig:first-3}
\end{subfigure}
\hfill
\begin{subfigure}{0.28\textwidth}
    \includegraphics[width=\textwidth]{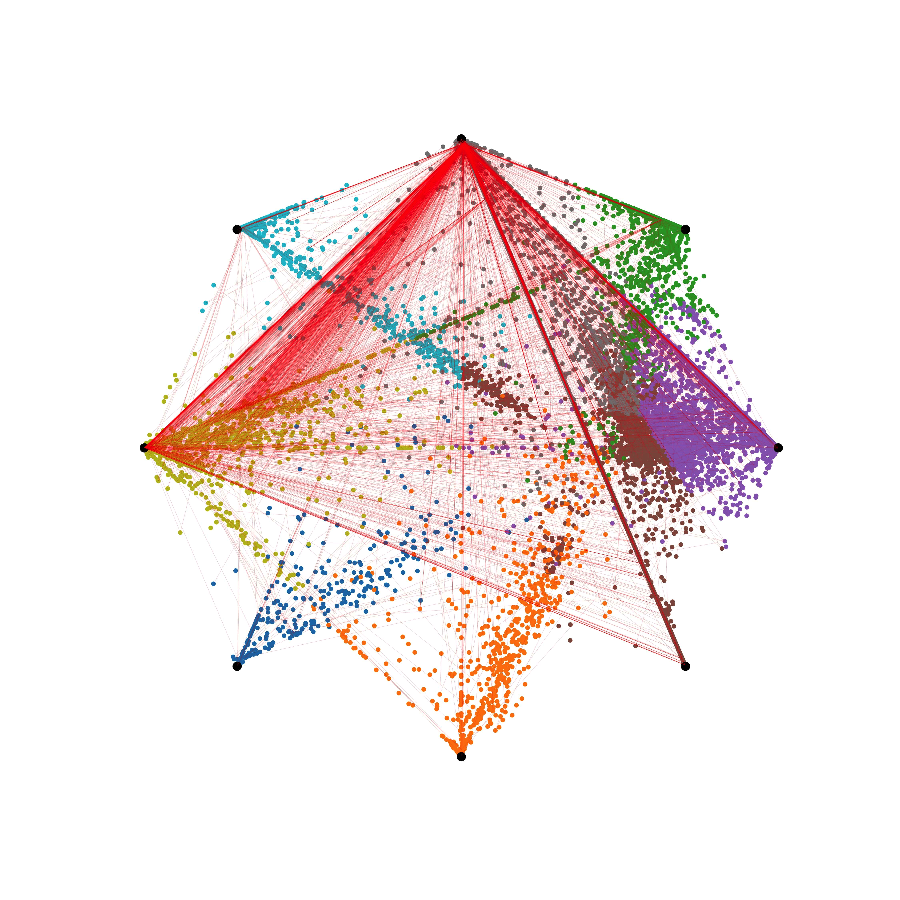}
    \caption{\textsl{Reddit}}
    \label{fig:second-3}
\end{subfigure}
\hfill
\begin{subfigure}{0.28\textwidth}
    \includegraphics[width=\textwidth]{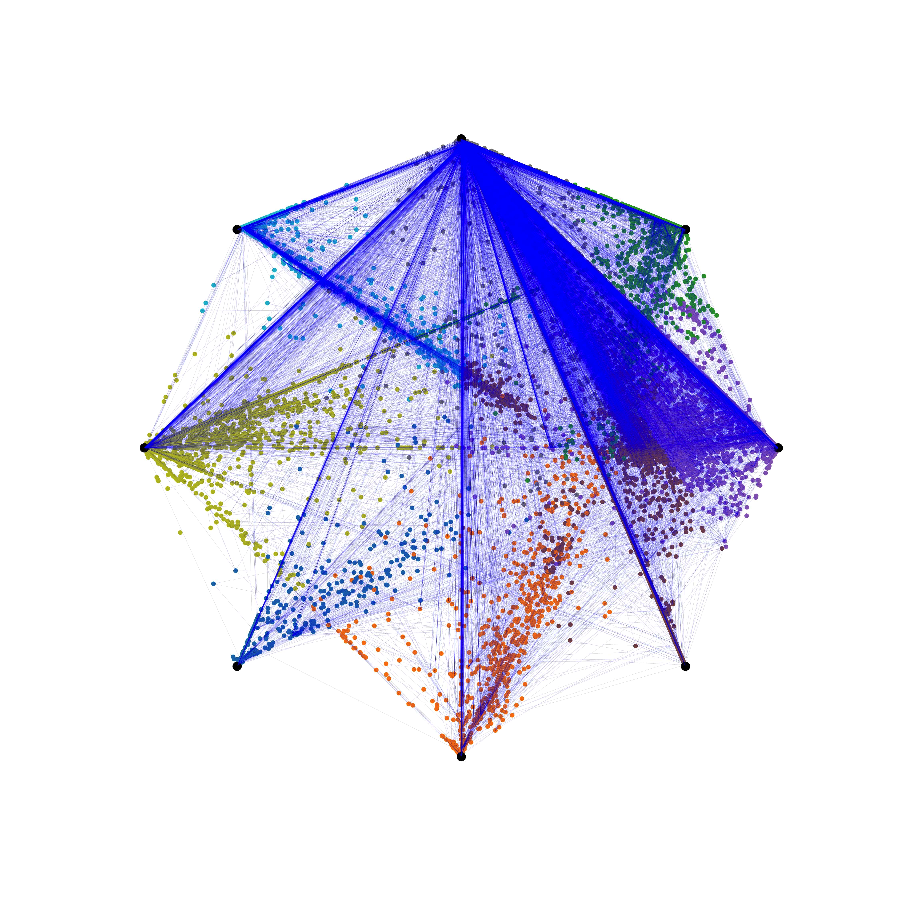}
    \caption{\textsl{Reddit}}
    \label{fig:third-3}
\end{subfigure}
\begin{subfigure}{0.28\textwidth}
    \includegraphics[width=\textwidth]{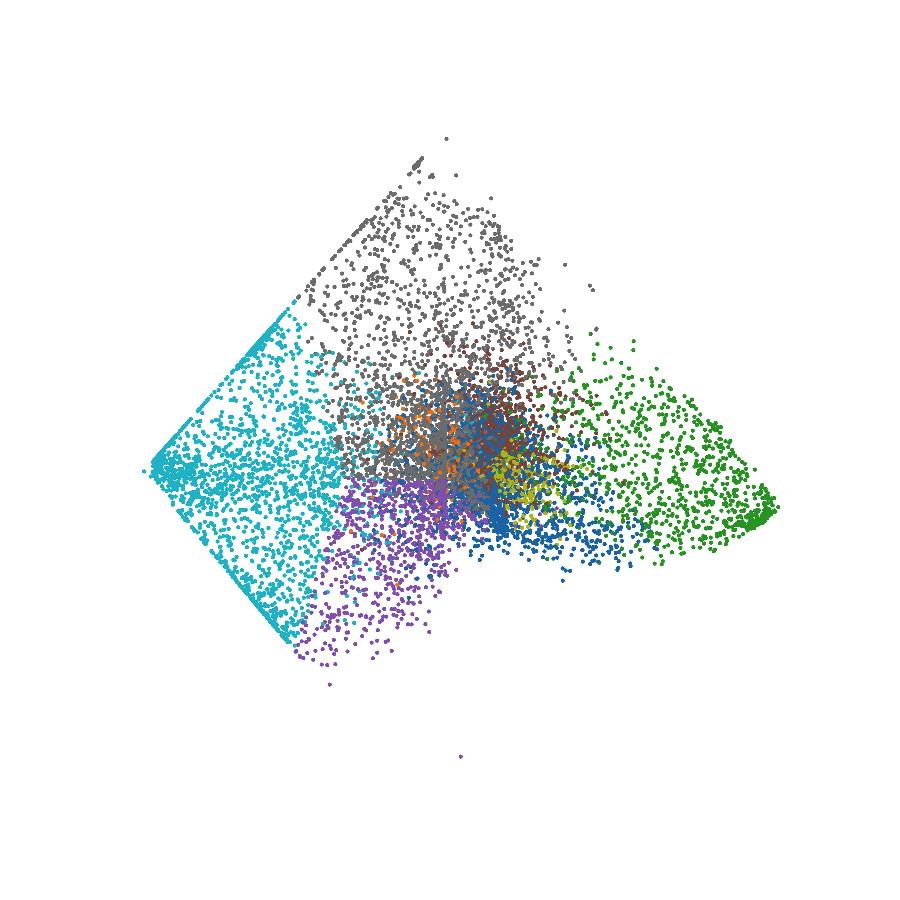}
    \caption{\textsl{Twitter}}
    \label{fig:first-4}
\end{subfigure}
\hfill
\begin{subfigure}{0.28\textwidth}
    \includegraphics[width=\textwidth]{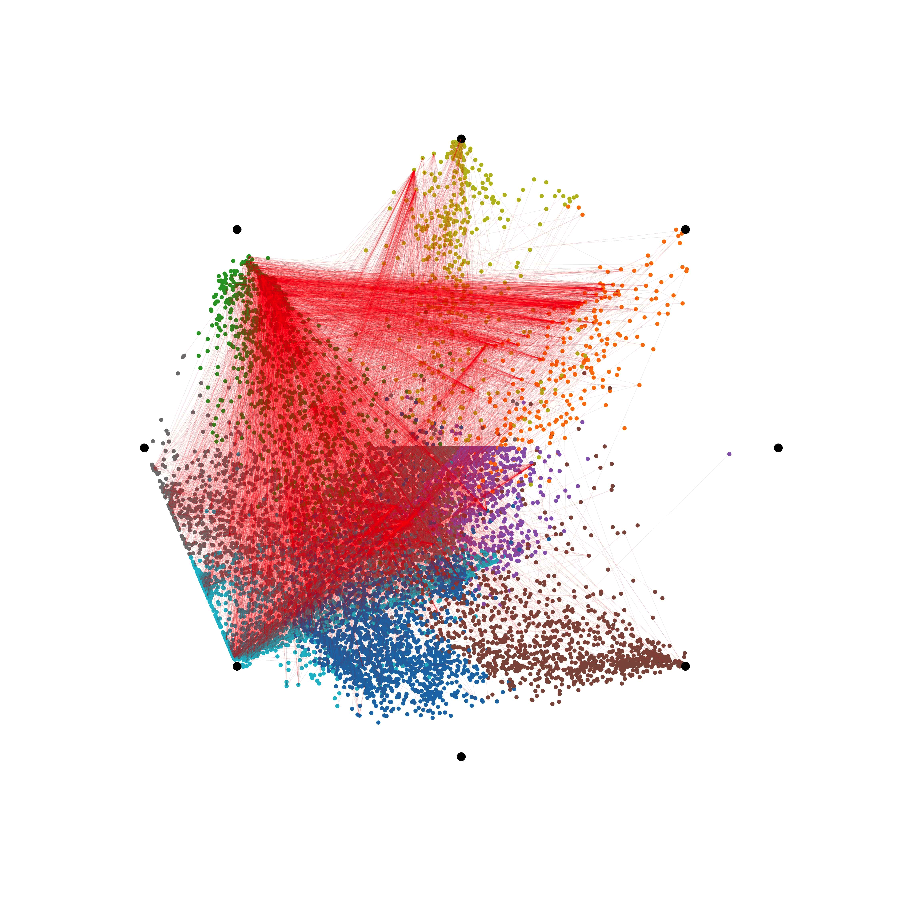}
    \caption{\textsl{Twitter}}
    \label{fig:second-5}
\end{subfigure}
\hfill
\begin{subfigure}{0.28\textwidth}
    \includegraphics[width=\textwidth]{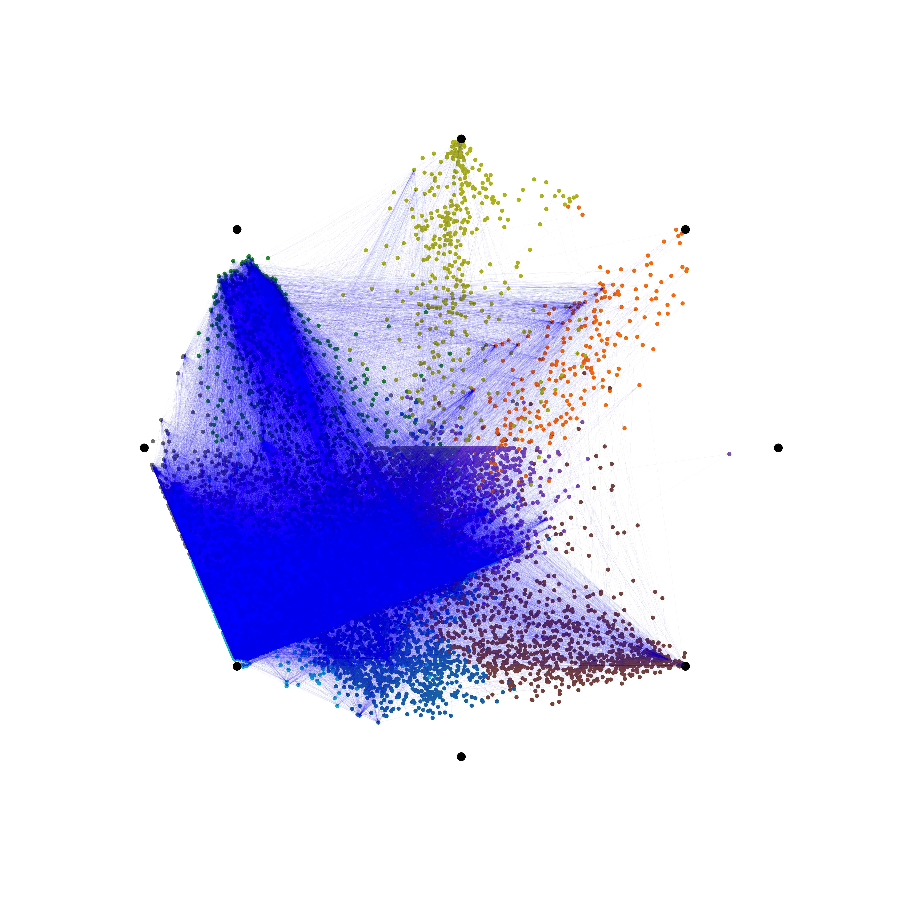}
    \caption{\textsl{Twitter}}
    \label{fig:third-6}
\end{subfigure}
\caption{Inferred polytope visualizations for various networks. The first column showcases the $K=8$ dimensional sociotope projected on the first two principal components (PCA) --- second and third columns provide circular plots of the sociotope enriched with the negative (red) and positive (blue) links, respectively.}
\label{fig:soc_viz}
\end{figure*}

\textbf{Visualizations.}
The RAA formulation facilitates the inference of a polytope describing the distinct aspects of networks. Here, we visualize the latent space across $K=8$ dimensions for all of the corresponding networks. To facilitate visualizations we use Principal Component Analysis (PCA), and project the space based on the first two principal components of the final embedding matrix $\Tilde{\bm{Z}}=\bm{AZ}$. In addition, we provide circular plots where each archetype of the polytope is mapped to a circle every $\text{rad}_k=\frac{2\pi}{K}$ radians, with $K$ being the number of archetypes. Figure \ref{fig:soc_viz} contains three columns with the first denoting the PCA-induced space while the second and third columns correspond to the circular plots enriched by the negative (red) and positive (blue) links, respectively. We observe how the polytope successfully uncovers extreme positional nodes. More specifically, all networks have at least one archetype which acts as a "dislike" hub and at least one as a "like" hub. Meaning that these archetypes contain high values of negative/positive interactions. For the \textsl{wiki-RfA} and \textsl{Twitter} networks we observe archetypes of very low degree, this is explained due to some only "disliked" nodes being pushed away from the main node population. These can be regarded as "outliers" of the sociotope. Nevertheless, such outliers are discovered since they provide high expressive power for the model.

\textbf{Discussion.} The Signed Relational Latent Distance Model has been presented for the undirected case setting, and we employed the Euclidean distance for both Skellam rates $\lambda^{+}_{ij},\, \lambda^{-}_{ij}$. The capacity of the current formulation works well for undirected networks. Nevertheless, there are alternative model formulations, and keeping the distance identical for the positive and negative rates constrains the models' expressive capability, especially for the directed/bipartite signed network case. We therefore explore additional model formulations such as setting the Skellam rates as, $\lambda_{ij}^+=\exp(\beta_i+\beta_j-||\mathbf{z}_i-\mathbf{w}_j||_2)$
and $\lambda_{ij}^-=\exp(\gamma_i+\gamma_j-||\boldsymbol{u}_i-\boldsymbol{w}_j||_2)$ in the supplementary material. Under this assumption, a positive directed relationship $(i \rightarrow j)$ shows that node $i$ "likes" node $j$ and "dislikes" node $j$ if it is negative. The latent embedding  $\boldsymbol{w}_j$ is then the receiver position for the "likes" and "dislikes" with embeddings $\mathbf{z}_i$ and $\boldsymbol{u}_i$ being the sender positions for positive and negative relationships, respectively. In this case, we introduce three latent embeddings instead of the conventional two for the undirected case. The disparity of location $\mathbf{z}_i$ and $\boldsymbol{u}_i$ here can point out how polarity is formed between the two regions of the latent space (Please see the supplementary material for further discussion and results).

Another important design characteristic for the \textsc{SLDM}/\textsc{SLIM} frameworks is the choice of the prior/regularization of the different parameters. So far, we did not tune any regularization strength of the priors and simply adopted a normal distribution on the model parameters and non-informative uniform Dirichlet prior on $\mathbf{Z}$ in the case of \textsc{SLIM}. Potential tuning of the priors with cross-validation is expected to boost the performance and results.

A prominent characteristic of signed networks is the sparsity or, in other words, the excess of "zero" weights among node pairs. An intriguing direction to account for it might be the zero-inflated version of the Skellam distribution \citep{skellam_sport}. Here essentially, we can define a mixture model responsible for the imbalance between cases (sign-weighted links) and controls (neutral zero links) in the network. Such zero-inflated \textsc{SLDM}/\textsc{SLIM} models can thereby define a generative process that can straightforwardly address different levels of network sparsity.

Whereas we consider the generalization of \textsc{SLDM} and \textsc{SLIM} to directed networks in the supplementary, a possible future direction should consider generalizations to bipartite networks in which we expect the directed generalizations to be applicable \citep{kim2018review,nakis2022a}. Furthermore, networks of polarization typically evolve over time. Future work should thus investigate how the proposed modeling framework can be extended to characterize dynamic networks leveraging existing works by exploring dynamic extensions of latent space modeling approaches, including the diffusion model of \citep{ldm_1} and approaches reviewed in \cite{kim2018review}.

%% file: 5-conclusion.tex
\section{CONCLUSION AND LIMITATIONS}\label{sec:conclusion}
The proposed Skellam Latent Distance Model (\textsc{SLDM}) and Signed Latent Relational Distance model (\textsc{SLIM}) provide easily interpretable network visualization with favorable performance in the link prediction tasks for weighted signed networks. In particular, endowing the model with a space constrained to polytopes (forming the \textsc{SLIM}) enabled us to characterize distinct aspects in terms of extreme positions in the social networks akin to conventional archetypal analysis but for graph-structured data. The Skellam distribution is considerably beneficial in modeling signed networks, whereas the relational extension of AA can be applied for other likelihood specifications, such as LDMs in general. This work thereby provides a foundation for using likelihoods accommodating weighted signed networks and representations akin to AA in general for analyzing networks.

The optimization for the \textsc{SLDM}/\textsc{SLIM} frameworks is a highly non-convex problem and thus relies on the quality of initialization in terms of convergence speed. In this regard, we use a deterministic initialization based on the normalized Laplacian. In addition, we observed that a maximum likelihood estimation of the model parameters became unstable when the network contained some nodes having only negative interactions. This is a direct consequence of the presence of the distance term ($\exp(+||\cdot||_2)$) for negative interactions, which can lead to overflow during inference. Nevertheless, the adopted MAP estimation was found to be stable across all networks. For real networks, the generative model created an "excess" of negative links increasing the overall network sparsity. For that, a modified \textsl{SLIM} excluding the regularization over the model parameters was introduced which achieved correct network sparsity (as shown in supplementary). Assuming priors over the model parameters created a bias over the generated network when compared to the ground truth network statistics.

%% file: supplement.tex
% \documentclass[twoside]{article}

% \usepackage{aistats2023}
% \usepackage{bm}
% \usepackage{booktabs}
% \usepackage{lscape}
% \usepackage{apalike}
% \usepackage{amsmath}
% \usepackage{amsfonts}
% \usepackage{graphicx}
% \usepackage{subcaption}
% \usepackage{natbib}
% \setcitestyle{numbers,compress,sort,round}
% \bibliographystyle{abbrvnat}
% \setcitestyle{authoryear,open={(},close={)}, aysep={,}} % If your paper is accepted, change the options for the package
% % aistats2023 as follows:
% %
% %\usepackage[accepted]{aistats2023}

% \begin{document}
\onecolumn
% \section{Supplementary Materials}
% \aistatstitle{Supplementary Materials}
% \onecolumn[\aistatstitle{Supplementary Materials}]
% \setcounter{section}{0}
\section{Directed Case Model Formulations}
In this section, we describe how our proposed frameworks can be extended to the study of directed networks, and we further explore additional model formulations allowing for more capacity and expressive power.

\subsection{The Skellam Latent Distance Model for the Directed Case (LDM)}
Our main purpose here is to learn two latent node representations $\{\mathbf{z}_i\}_{i\in\mathcal{V}}\in\mathbb{R}^{K}$ and $\{\mathbf{w}_i\}_{i\in\mathcal{V}}\in\mathbb{R}^{K}$ in a low dimensional space for a given directed signed network $\mathcal{G}=(\mathcal{V}, \mathcal{Y})$ ($K \ll |\mathcal{V}|$). The two sets of the latent embeddings correspond to modeling directed relationships $i\rightarrow j$ of nodes, with $\mathbf{z}_i$ the source node and $\mathbf{w}_j$ the target node, and vice-versa for an oppositely directed relationship $i\leftarrow j$. Similar to the main paper, we can formulate the negative log-likelihood of a latent distance model under the Skellam distribution as:
\begin{align*}
\mathcal{L}(\mathcal{Y}) &:=\log p(y_{ij}|\lambda^{+}_{ij},\lambda^{-}_{ij}) \\
&= \sum_{i,j}{(\lambda^{+}_{ij}+\lambda^{-}_{ij})} - \frac{y_{ij}}{2}\log\left(\frac{\lambda^{+}_{ij}}{\lambda^{-}_{ij}}\right)-\log\left(\mathcal{I}_{|y_{ij}|}\left(2\sqrt{\lambda^{+}_{ij}\lambda^{-}_{ij}}\right)\right),
\end{align*}
For the directed case, the Skellam distribution has two rate parameters as well, and we consider them to learn latent node representations $\{\mathbf{z}_i\}_{i\in\mathcal{V}}$ and $\{\mathbf{w}_j\}_{j\in\mathcal{V}}\in\mathbb{R}^{K}$ by defining them as follows:
\begin{align}
\lambda_{ij}^{+} &= \exp\big(\beta_{i} + \gamma_{j} - ||\mathbf{z}_i-\mathbf{w}_j||_2\big)\label{eq:rate1},
\\
\lambda_{ij}^{-} &= \exp\big(\delta_{i} + \epsilon_{j} + ||\mathbf{z}_i-\mathbf{w}_j||_2\big),
\label{eq:rate21}
\end{align}
where the set $\{\beta_{i},\gamma_i,\delta_i,\epsilon_i\}_{i\in\mathcal{V}}$ denote the node-specific random effect terms, and $||\cdot||_2$ is the Euclidean distance function. More specifically, the sender $\beta_i$ and the receiver $\gamma_j$ random effects represent the "social" reach of a node and the tendency to form positive interactions, expressing positive degree heterogeneity (indicated by $+$ as a superscript of $\lambda$). In contrast, $\delta_i$ and $\epsilon_j$ provide the "anti-social" sender and receiver effect of a node to form negative connections, and thus model negative degree heterogeneity (indicated by $-$ as a superscript of $\lambda$). 

By imposing (as in the undirected case) standard normally distributed priors elementwise on all model parameters $\bm{\theta}=\{\bm{\beta},\boldsymbol{\gamma},\bm{\delta},\bm{\epsilon}, \mathbf{Z},\boldsymbol{W}\}$, i.e., $\theta_i\sim \mathcal{N}(0,1)$, We define a maximum a posteriori (MAP) estimation over the model parameters, via the loss function to be minimized (ignoring constant terms):
\begin{align}\label{eq:loss_sk}
\begin{split}
    Loss =& \sum_{i,j}\Bigg( \lambda_{ij}^{+}+\lambda_{ij}^{-} -\frac{y_{ij}}{2}\log\left( \frac{\lambda_{ij}^{+} }{\lambda_{ij}^{-}}\right)\Bigg) - \sum_{i,j}\log I_{|y_{ij}|}\Big (2\sqrt{\lambda_{ij}^{+}\lambda_{ij}^{-}}\Big )
    \\
    +& \frac{\rho}{2}\Big(||\mathbf{Z}||_F^2+||\mathbf{W}||_F^2+||\bm{\gamma}||_F^2+||\bm{\beta}||_F^2+||\bm{\delta}||_F^2+||\bm{\epsilon}||_F^2\Big),
    \end{split}
\end{align}
where $||\cdot||_F$ denotes the Frobenius norm. In addition, $\rho$ is the regularization strength with $\rho=1$ yielding the adopted normal prior with zero mean and unit variance. 

\subsection{The Signed Relational Latent Distance Model for Directed Networks}
We formulate the relational AA in the context of the family of LDMs and for directed networks, as:
\begin{align}
    \lambda_{ij}^{+} &=\exp \big( \beta_{i} + \gamma_{j} - \|\mathbf{A} (\mathbf{z}_{i}-\mathbf{w}_{j})\|_{2}\big)
    \\ 
    &=\exp \big( \beta_{i} + \gamma_{j} -\|\mathbf{R}[\mathbf{Z};\mathbf{W}]\mathbf{C}(\mathbf{z}_{i}-\mathbf{w}_{j})\|_2\big).\label{LRPM_inensity_function_1}
    \\
\lambda_{ij}^{-} &=\exp \big( \delta_{i} + \epsilon_{j} + \|\mathbf{A} (\mathbf{z}_{i}-\mathbf{w}_{j})\|_{2}\big)
\\ 
&=\exp \big( \delta_{i} + \epsilon_{j} +\|\mathbf{R}[\mathbf{Z};\mathbf{W}]\mathbf{C}(\mathbf{z}_{i}-\mathbf{w}_{j})\|_2\big).\label{LRPM_inensity_function_2}
\end{align}

Notably, in the AA formulation $\mathbf{X}=\mathbf{R}[\mathbf{Z};\mathbf{W}]$ corresponds to observations formed by the concatenations of the convex combinations $\mathbf{Z}$ and $\boldsymbol{W}$ of positions given by the columns of $\boldsymbol{R}^{K\times K}$. Furthermore, in order to ensure what is used to define archetypes $\mathbf{A}=\mathbf{XC}=\mathbf{R}[\mathbf{Z};\mathbf{W}]\mathbf{C}$ corresponds to observations using these archetypes in their reconstruction $[\mathbf{Z};\mathbf{W}]$,
we define $\boldsymbol{C}\in \boldsymbol{R}^{2N\times K}$ as a gated version of $[\mathbf{Z};\mathbf{W}]$ normalized to the simplex such that $\boldsymbol{c}_d\in\Delta^{2N}$ by defining 
\begin{equation}
    c_{nd}=\frac{([\mathbf{Z};\mathbf{W}]^\top\circ [\sigma(\mathbf{G})]^\top)_{nd}}{\sum_{n^\prime}([\mathbf{Z};\mathbf{W}]^\top\circ [\sigma(\mathbf{G})]^\top)_{n^\prime d}}
\end{equation}
in which $\circ$ denotes the elementwise (Hadamard) product and $\sigma(\mathbf{G})$ defines the logistic sigmoid elementwise applied to the matrix $\boldsymbol{G}$. As a result, the extracted archetypes are ensured to correspond to the nodes assigned the archetype, whereas the location of the archetypes can be flexibly placed in space as defined by $\mathbf{R}$. By defining $\mathbf{z}_i=\operatorname{softmax}(\tilde{\mathbf{z}}_i)$ and $\boldsymbol{w}_i=\operatorname{softmax}(\tilde{\boldsymbol{w}}_i)$ we further ensure $\mathbf{z}_i,\boldsymbol{w}_i\in \Delta^K$.
% We further assume node embeddings $\{\mathbf{z}\}_{i\in\mathcal{V}}$ lie in the standard simplex set, $\Delta^K$.
%Thus, the product $\mathbf{ZCz}_{i}$ corresponds to the archetypal convex projection of the latent embedding of the ith node.-\textbf{Added by Christian (maybe redundant)} 

As in the undirected case, the loss function of Eq. \eqref{eq:loss_sk} is adopted for the relational AA formulation forming the \textsc{SLIM}, with the prior regularization applied to the corners of the extracted polytope $\bm{A}=\mathbf{R}[\mathbf{Z};\mathbf{W}]\mathbf{C}$ instead of the latent embeddings $\bm{Z},\bm{W}$ imposing a standard elementwise normal distribution as prior $a_{k,k^\prime}\sim \mathcal{N}(0,1)$. Furthermore, we impose a uniform Dirichlet prior on the columns of $\bm{Z},\bm{W}$, i.e. $(\bm{z}_i,\bm{w}_i\sim Dir(\mathbf{1}_K)$, this only contributes constant terms to the joint distribution. As a result, the loss function is given by Eq. \eqref{eq:loss_sk} replacing $\|\mathbf{Z}\|_F^2$ and $\|\boldsymbol{W}\|_F^2$ with $\|\boldsymbol{A}\|_F^2$ for the maximum a posteriori (MAP) optimization.

\subsection{Model Extensions for Additional Capacity}\label{model_extension}
In the main paper, we briefly introduced an additional formulation for the rates of the Skellam distribution as adopted by our models. In this case (and for directed networks), the rates are:
\begin{align}
    \lambda_{ij}^+=\exp(\beta_i+\gamma_j-||\mathbf{z}_i-\mathbf{w}_j||_2)
\ \text{and} \
% \end{equation}
% \begin{equation}
    \lambda_{ij}^-=\exp(\delta_i+\epsilon_j-||\mathbf{u}_i-\mathbf{w}_j||_2)
\end{align}
 In this proposition, we have adopted three latent embeddings instead of the two previously described for the directed case. The disparity of location $\mathbf{z}_i$ and $\mathbf{u}_i$ here can point out how polarity is formed between the two regions of the latent space. This model specification introduces an additional regularization for the third embedding matrix $\mathbf{U}$ in the loss function of Equation \eqref{eq:loss_sk}. For the RAA case, we thereby define $\mathbf{X}=\mathbf{R}[\mathbf{Z};\mathbf{U};\mathbf{W}]$, i.e., as the concatenation of all three latent positions and with $\boldsymbol{C}\in \boldsymbol{R}^{3N\times K}$.
 
 \subsection{Directed case --- Results and performance}
\begin{figure*}[!h]
\centering
\begin{subfigure}{0.24\textwidth}
    \includegraphics[width=\textwidth]{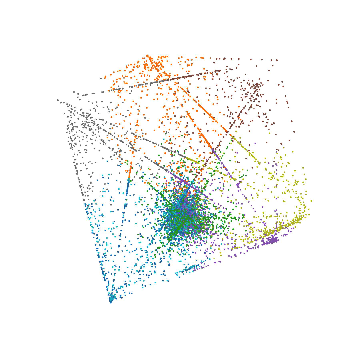}
    \caption{\textsl{WikiElec}}
\end{subfigure}
\hspace{1cm}%\hfill
\begin{subfigure}{0.24\textwidth}
    \includegraphics[width=\textwidth]{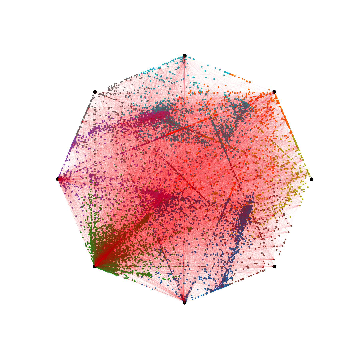}
    \caption{\textsl{WikiElec}}
\end{subfigure}
\hspace{1cm}%\hfill
\begin{subfigure}{0.24\textwidth}
    \includegraphics[width=\textwidth]{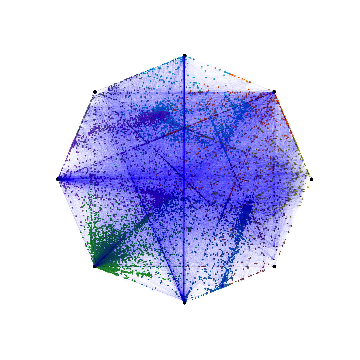}
    \caption{\textsl{WikiElec}}
\end{subfigure}
\newline
\begin{subfigure}{0.24\textwidth}
    \includegraphics[width=\textwidth]{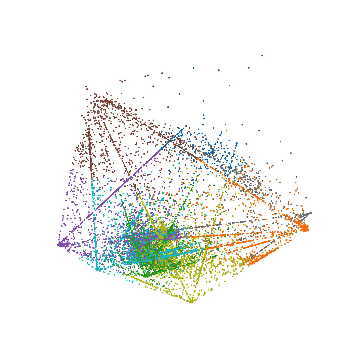}
    \caption{\textsl{WikiRfa}}
\end{subfigure}
\hspace{1cm}%\hfill
\begin{subfigure}{0.24\textwidth}
    \includegraphics[width=\textwidth]{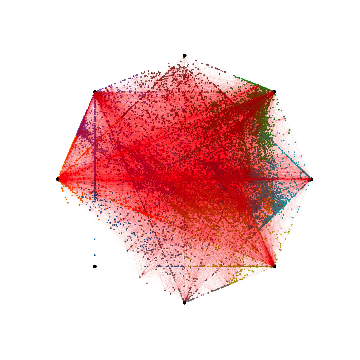}
    \caption{\textsl{WikiRfa}}
\end{subfigure}
\hspace{1cm}%\hfill
\begin{subfigure}{0.24\textwidth}
    \includegraphics[width=\textwidth]{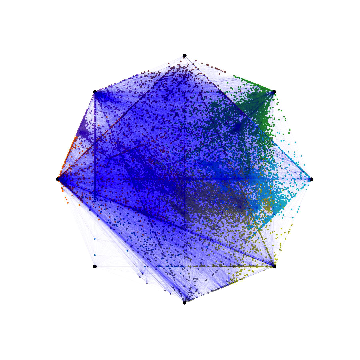}
    \caption{\textsl{WikiRfa}}
\end{subfigure}
\newline
\begin{subfigure}{0.24\textwidth}
    \includegraphics[width=\textwidth]{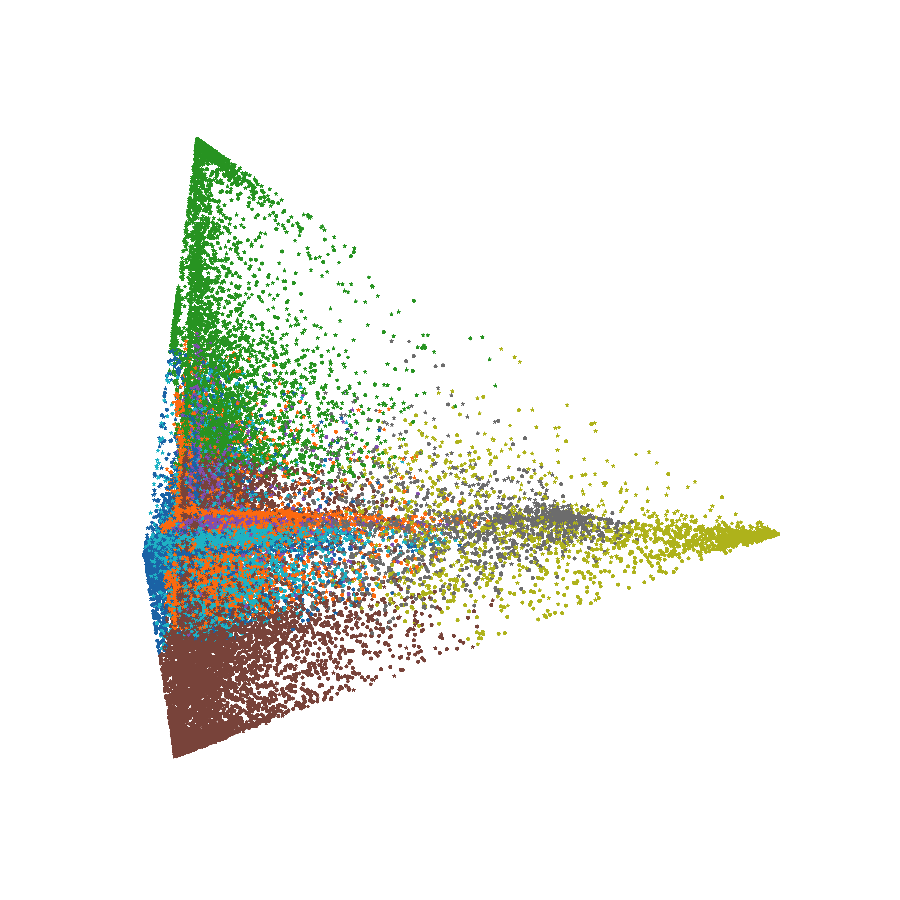}
    \caption{\textsl{Reddit}}
\end{subfigure}
\hspace{1cm}%\hfill
\begin{subfigure}{0.24\textwidth}
    \includegraphics[width=\textwidth]{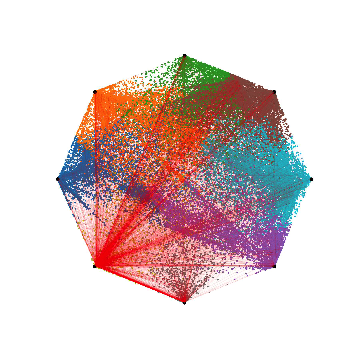}
    \caption{\textsl{Reddit}}
\end{subfigure}
\hspace{1cm}%\hfill
\begin{subfigure}{0.24\textwidth}
    \includegraphics[width=\textwidth]{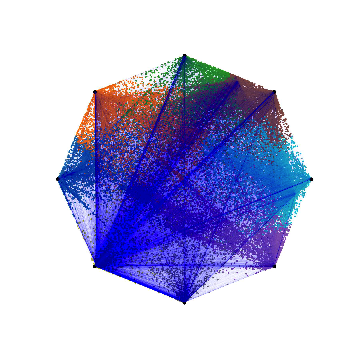}
    \caption{\textsl{Reddit}}
\end{subfigure}
\caption{Inferred polytope visualizations for various directed networks. The first column showcases the $K=8$ dimensional sociotope projected on the first two principal components (PCA) of the combined embeddings (source, target) $[\mathbf{Z};\mathbf{W}]$ --- second and third columns provide circular plots of the sociotope enriched with the negative (red) and positive (blue) links, respectively.}
\label{fig:soc_viz1}
\end{figure*}

 In Table \ref{tab:auc_roc1} and Table \ref{tab:auc_pr1}, we provide the results for the directed networks against various prominent baselines. Note that \textsc{POLE} is not defined for the directed case while \textsc{SIDE} failed to create embeddings for one-degree nodes. For the frameworks, we use two additional variations for \textsc{SLDM} and \textsc{SLIM}. The first ones are the \textsc{SLDM reg=0.01} and \textsc{SLIM reg=0.01}, where we have used a regularization power $\rho=0.01$ in Equation \eqref{eq:loss_sk}. This shows how performance is affected by less regularized parameters. In addition, we also provide results for \textsc{SLDM-expr} and \textsc{SLIM-expr} which denote the more expressive model as described in Subsection \ref{model_extension}. The results showcase our models' capability to outperform the baselines or provide competitive performance. Comparing now the \textsc{SLDM} and \textsc{SLIM} different variations we observe that performance is boosted by just using the vanilla methods. It seems that the most important trait is the regularization power of the model rather than the expressive capabilities that extra parameters provide to the model. Lastly, Figure \ref{fig:soc_viz1} provides the same visualizations as in the main paper but for the directed networks.

\begin{table*}[!b]
\centering
\caption{Area Under Curve (AUC-ROC) scores for varying representation sizes (Directed). The symbol '-' denotes that the corresponding model is not able to run on directed networks while 'x' that the model returned errors.}
\label{tab:auc_roc1}
\resizebox{0.8\textwidth}{!}{%
\begin{tabular}{rccccccccccccccccccccccccc}\toprule
\multicolumn{1}{l}{} & \multicolumn{3}{c}{\textsl{WikiElec}} & \multicolumn{3}{c}{\textsl{WikiRfa}}& \multicolumn{3}{c}{\textsl{Reddit}} \\\cmidrule(rl){2-4}\cmidrule(rl){5-7}\cmidrule(rl){8-10}
\multicolumn{1}{r}{Task} & $p@n$ & $p@z$ & $n@z$ & $p@n$ & $p@z$ & $n@z$ & $p@n$ & $p@z$ & $n@z$  \\\cmidrule(rl){1-1}\cmidrule(rl){2-2}\cmidrule(rl){3-3}\cmidrule(rl){4-4}\cmidrule(rl){5-5}\cmidrule(rl){6-6}\cmidrule(rl){7-7}\cmidrule(rl){8-8}\cmidrule(rl){9-9}\cmidrule(rl){10-10}
\textsc{POLE}  &- &- &- &- &- &- &- &- &- \\
\textsc{SLF}   &.938 &.971 &.980 &.991 &.980 &.985 &.823 &.974 &.984\\
\textsc{SiGAT}  &.921 &.750 &.871 &.988 &.772 &.927 &.982 &.713 &.980\\
\textsc{SIDE}     &x &x &x &x &x &x &x &x &x\\
\textsc{SigNet}     &.929 &.907 &.835 &.991 &.921 &.873 &.881 &.757 &.719 \\
\midrule
\textsc{SLIM  (ours)}     &.910 &.981 &.963 &.984 &.989 &.981 &.713 &.973 &.982 \\    
\textsc{SLDM (ours)}    &.914 &.977 &.966 &.983 &.987 &.978 &.657 &.937 &.964\\
\textsc{SLIM reg=0.01 (ours)}    &.927 &.989 &.980 &.992 &.994 &.990 &.827 &.982 &.989 \\    
\textsc{SLDM reg=0.01 (ours)}   &.940 &.989 &.980 &.984 &.987 &.976 &.774 &.982 &.986\\
\textsc{SLIM-expr (ours)}    &.922 &.984 &.977 &.987 &.988 &.982 &.706 &.930 &.949 \\
\textsc{SLDM-expr (ours)}    &.915 &.987 &.985 &.989 &.994 &.992 &.657 &.965 &.965 \\
\bottomrule 
\end{tabular}}
\end{table*}
\begin{table*}[!t]
\centering
\caption{Area Under Curve (AUC-PR) scores for varying representation sizes (Directed). The symbol '-' denotes that the corresponding model is not able to run on directed networks while 'x' that the model returned errors.}
\label{tab:auc_pr1}
\resizebox{0.8\textwidth}{!}{%
\begin{tabular}{rccccccccccccccccccccccccc}\toprule
\multicolumn{1}{l}{}  & \multicolumn{3}{c}{\textsl{WikiElec}} & \multicolumn{3}{c}{\textsl{WikiRfa}} & \multicolumn{3}{c}{Reddit} \\\cmidrule(rl){2-4}\cmidrule(rl){5-7}\cmidrule(rl){8-10}
\multicolumn{1}{r}{Task} & $p@n$ & $p@z$ & $n@z$ & $p@n$ & $p@z$ & $n@z$ & $p@n$ & $p@z$ & $n@z$  \\\cmidrule(rl){1-1}\cmidrule(rl){2-2}\cmidrule(rl){3-3}\cmidrule(rl){4-4}\cmidrule(rl){5-5}\cmidrule(rl){6-6}\cmidrule(rl){7-7}\cmidrule(rl){8-8}\cmidrule(rl){9-9}\cmidrule(rl){10-10}
\textsc{POLE}     & - & - & - & - & - & - & - & -  & -\\
\textsc{SLF}     &.981 &.949 &.890 &.995 &.954 &.951 &.978 &.972 &.919\\
\textsc{SiGAT}    &.977 &.689 &.562 &.993 &.685 &.714 &.998 &.727 &.659 \\
\textsc{SIDE}     &x &x &x &x &x &x &x &x &x\\
\textsc{SigNet}    &.979 &.831 &.577 &.995 &.840 &.671 &.988 &.675 &.233 \\
\midrule
\textsc{SLIM (ours)}     &.971 &.974 &.852 &.989 &.981 &.951 &.962 &.971 &.874\\
\textsc{SLDM  (ours)}    &.972 &.967 &.862 &.988 &.978 &.939 &.952 &.948 &.861 \\
\textsc{SLIM reg=0.01 (ours)}    &.976 &.983 &.910 &.995 &.988 &.973 &.980 &.982 &.918 \\
\textsc{SLDM  reg=0.01 (ours)}    &.981 &.983 &.912 &.991 &.976 &.930 &.972 &.981 &.911 \\
\textsc{SLIM-expr (ours)}     &.976 &.978 &.914 &.992 &.980 &.953 &.958 &.938 &.823 \\
\textsc{SLDM-expr (ours)}    &.973 &.981 &.936 &.993 &.987 &.979 &.949 &.966 &.871 \\
\bottomrule    
\end{tabular}%
}
\end{table*}

\section{Initialization}
For the \textsc{SLDM} model, we used the Eigen-decomposition of the normalized Laplacian for singed networks \citep{norm_lapl}. Solving the eigenproblem for a few eigenvalues can be done efficiently through the Lanczos method \citep{10.5555/248979}, due to the high sparsity of real large-scale networks.

For \textsc{SLIM}, we would like to initialize matrix $\mathbf{A}$ based on the convex hull of the spectral decomposition of the normalized Laplacian. This is very costly since finding the convex hull has an exponential increase in complexity in terms of the dimensionality of the space. For that purpose, we use the furthest sum algorithm \citep{5589222} to discover guaranteed distinct aspects of the spectral space. Lastly, since we are unable to directly initialize $\mathbf{A}$, we use the furthest sum discovered points to initialize $\mathbf{R}$ while also tuning $\mathbf{G}$ for picking up the correct points in the latent space.

\section{Bessel Function Approximation}
We need to compute the modified Bessel function of the first kind and of order $y$ for the implementation of our proposed approach, which is defined by 
\begin{align*}
    I_{y}(x)=\big(\frac{x}{2} \big)^y\sum_{k=0}^{\infty}\frac{(\frac{x^2}{4})^k}{k!\Gamma(y+k+1)}
\end{align*}
We approximate the actual value by only considering the first $50$ terms of the infinite sum. Since we have small orders of $y$ and small values of $x$, the series components converge to zero quickly. We observed that taking the first $50$ components does not affect the performance/accuracy of the model.

\section{Generating based on real networks}
\begin{figure*}[!h]
\centering
\begin{subfigure}{.28\textwidth}
  \centering
  \includegraphics[width=\textwidth]{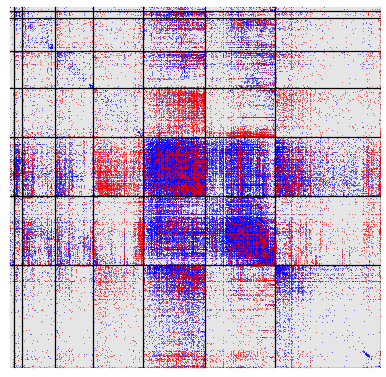}
  \caption{Ground Truth: $(.003,78\%,22\%)$}
\end{subfigure}
% \hfill
\begin{subfigure}{.28\textwidth}
  \centering
  \includegraphics[width=\textwidth]{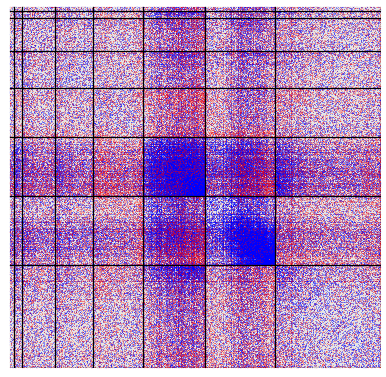}
  \caption{Generated: $(.006,63\%,37\%)$}
\end{subfigure}
\caption{\textsl{wikiElec} ground truth (left) adjacency matrix and generated (right) adjacency matrix based on inferred parameters. 
The parenthesis shows the network statistics as: (density,\% of positive (blue) links,\% of negative (red) links). All network adjacency matrices are ordered based on $\mathbf{z}_i$, in terms of maximum archetype membership and internally according to the magnitude of the corresponding archetype most used for their reconstruction.}
\label{fig:real_gen}
\end{figure*}

\begin{figure*}[!h]
\centering
\begin{subfigure}{.28\textwidth}
  \centering
  \includegraphics[width=\textwidth]{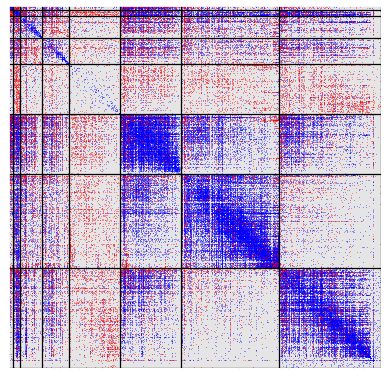}
  \caption{Ground Truth: $(.003,78\%,22\%)$}
\end{subfigure}
% \hfill
\begin{subfigure}{.28\textwidth}
  \centering
  \includegraphics[width=\textwidth]{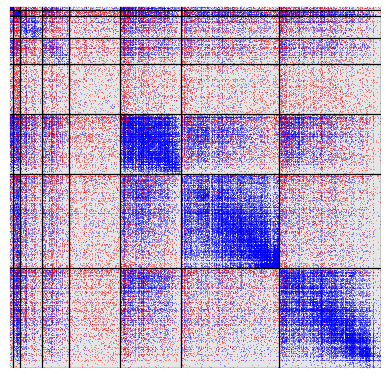}
  \caption{Generated: $(.003,76\%,24\%)$}
\end{subfigure}
\caption{\textsl{wikiElec} ground truth (left) adjacency matrix and generated (right) adjacency matrix based on inferred parameters with a \textsc{SLIM} \textbf{without regularization priors} over the parameters. 
The parenthesis shows the network statistics as: (density,\% of positive (blue) links,\% of negative (red) links). All network adjacency matrices are ordered based on $\mathbf{z}_i$, in terms of maximum archetype membership and internally according to the magnitude of the corresponding archetype most used for their reconstruction.}
\label{fig:real_gen_noreg}
\end{figure*}

Here, we test how the model generates based on real networks. We use the \textsl{wikiElec} to train an $K=8$ dimensional \textsc{SLIM} model and we then generate a network based on the inferred parameters. Results are shown in Fig. \ref{fig:real_gen} where we observe that the generated network learns successfully the main structure of the network but generates more non-zero elements and more negative links thereby decreasing the sparsity and increasing the percentage of negative links when compared to the ground truth. Modifying \textsl{SLIM} to exclude the regularization over the model parameters achieves correct network sparsity as shown in Fig \ref{fig:real_gen_noreg} with only a $2\%$ increase in the inferred percentage of negative links when compared to the ground truth. Adding priors to the model creates a bias over the network generation. Lastly, the un-regularized \textsc{SLIM} boosted performance in the link prediction tasks ranging from $1\%$ to $5\%$ for the \textsl{wikiElec} network. Nevertheless, priors over the model parameters stabilize the inference when "extreme" negative nodes exist in the network (nodes with only negative links) that can also be considered outliers.

 \section{Effect of Sampling Size}
\begin{figure}[!ht]
\centering
\begin{subfigure}{0.28\textwidth}
\centering
    \includegraphics[width=\textwidth]{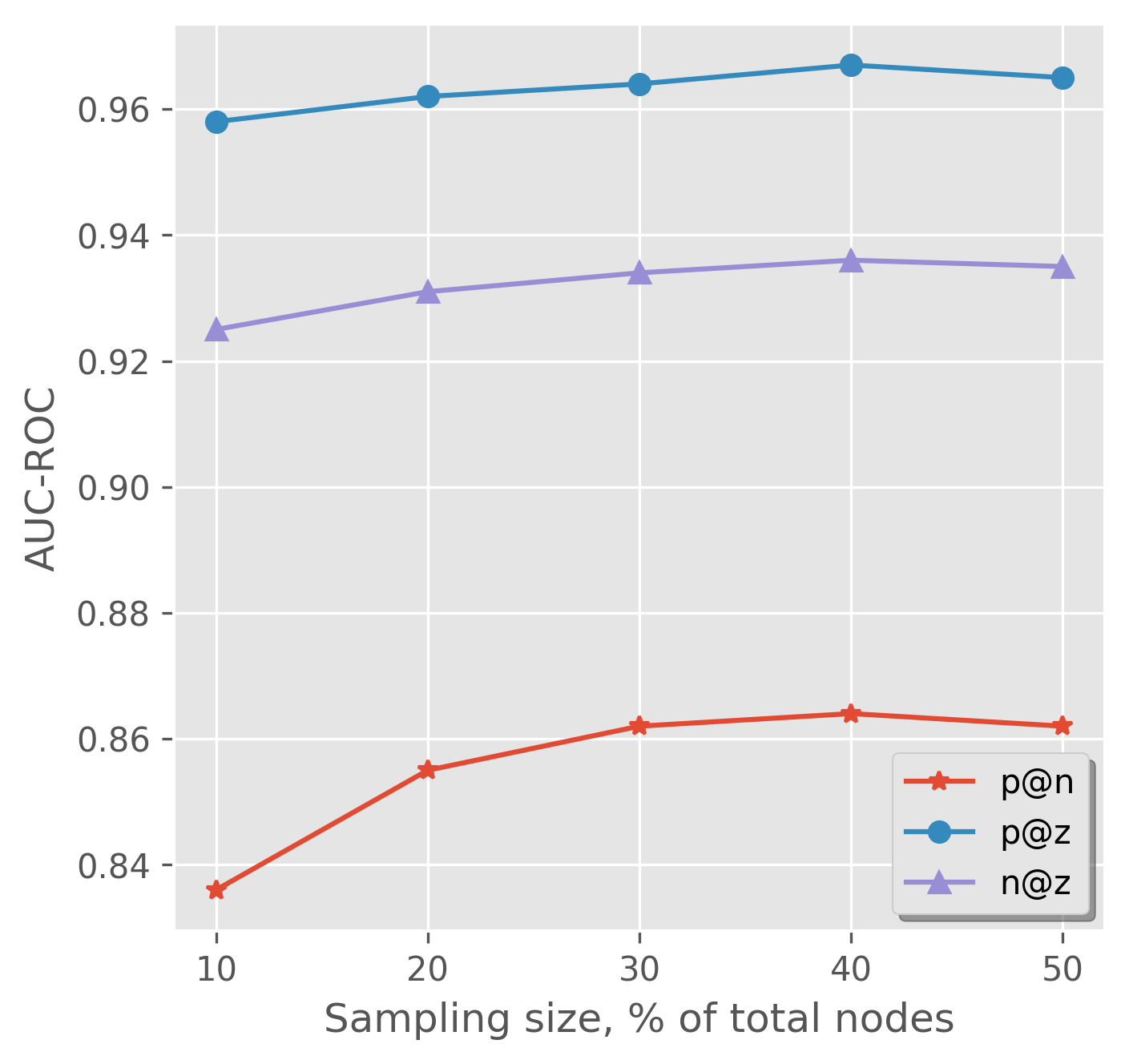}
    \caption{}
\end{subfigure}
% \hfill
\begin{subfigure}{0.28\textwidth}
\centering
    \includegraphics[width=\textwidth]{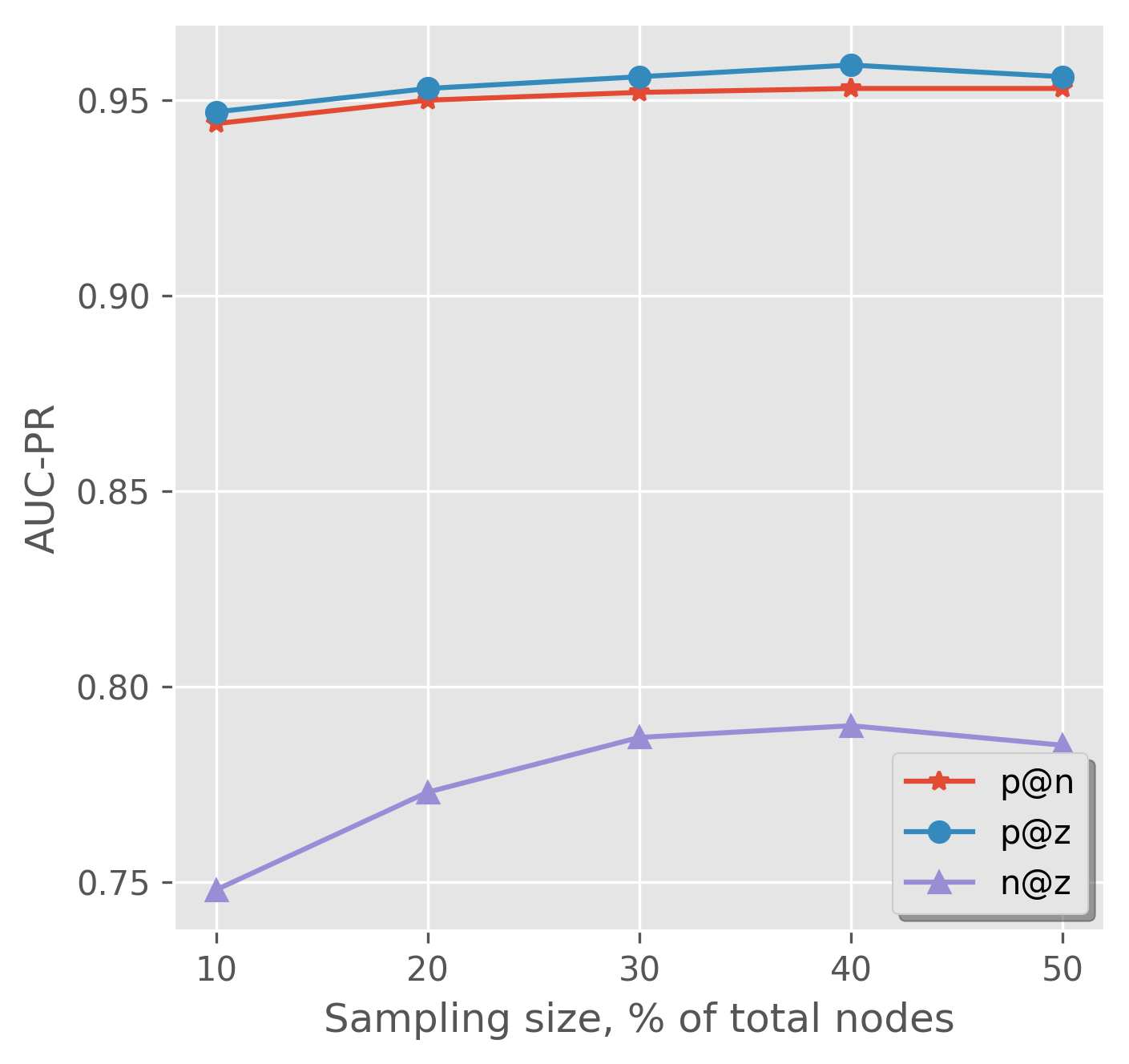}
        \caption{}
\end{subfigure}
\caption{\textsl{wikiElec}: Performance of \textsc{SLIM} across sample sizes for different tasks, (a) Area-Under-Curve Receiver Operating Characteristic scores, (b) Area-Under-Curve Precision-Recall scores. Both AUC-ROC and AUC-PR scores are almost constant across different dimensions }
\label{fig:sample_roc_pr1}
\end{figure}

In the main paper, the sample size was set to the maximum number ($\sim 3000$ nodes) that our $8$GB GPU could fit in memory. Here, we provide a study on how different sample sizes affect the performance of the \textsc{SLIM} model. In Fig \ref{fig:sample_roc_pr1}, we provide the performance across different tasks for the \textsl{wikiElec} dataset, considering sampling size of $\{10\%,20\%,30\%,40\%,50\%\}$. We observe with small differences almost constant performance across different sampling sizes. As we decrease the sampling size to $10\%$ and $20\%$ we observe some more significant decreases in the $p@ n$ task performance. This is because we keep the total number of training iterations (it=$5000$) for all cases. Overall, smaller sampling sizes require additional iterations to converge to the performance of the model with larger sampling sizes.

\section{Effect of Learning Rate}
\begin{figure*}[!ht]
\centering
\begin{subfigure}{0.28\textwidth}
    \centering
    \includegraphics[width=\textwidth]{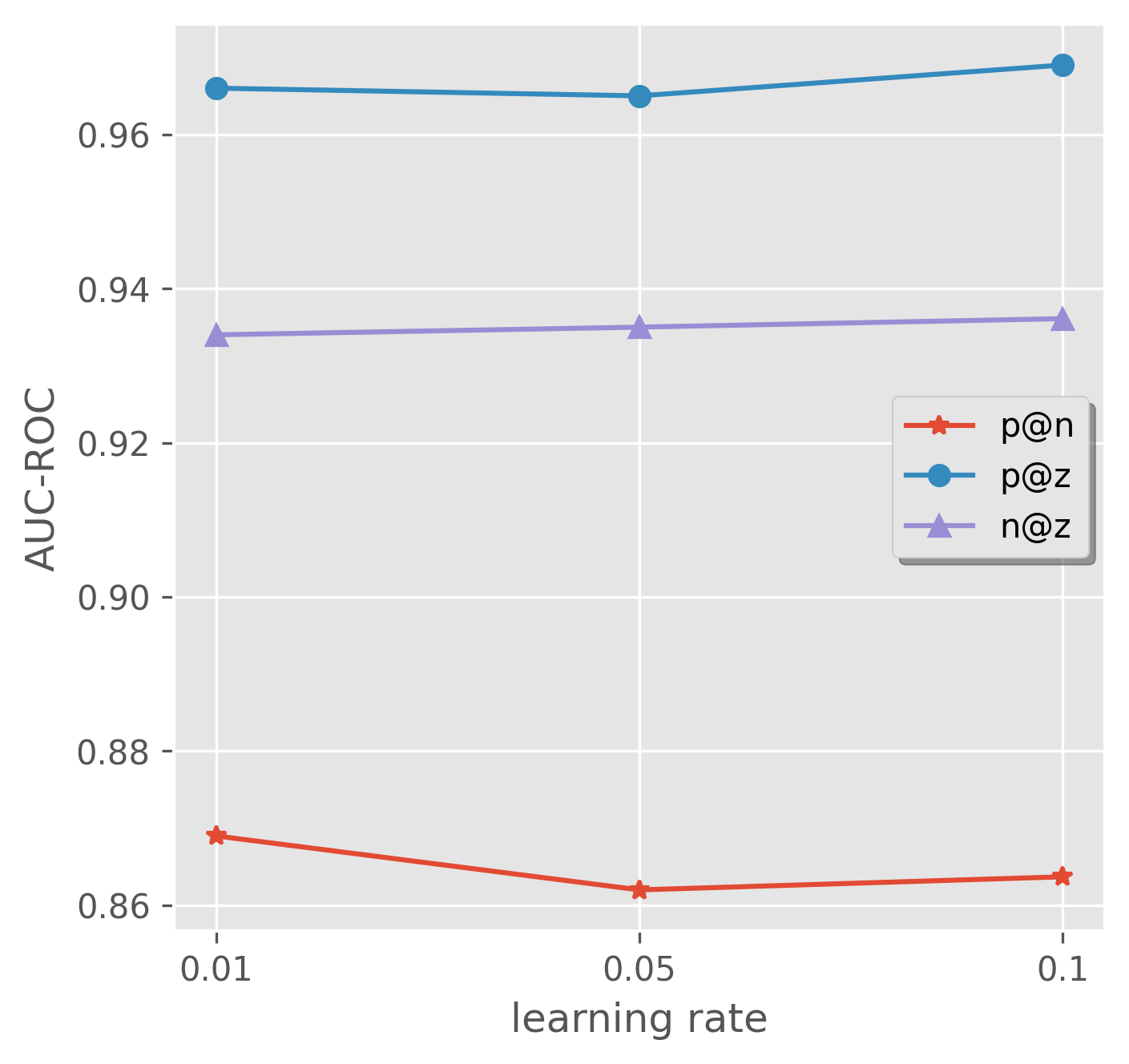}
    \caption{}
\end{subfigure}
% \hfill
\begin{subfigure}{0.28\textwidth}
\centering
    \includegraphics[width=\textwidth]{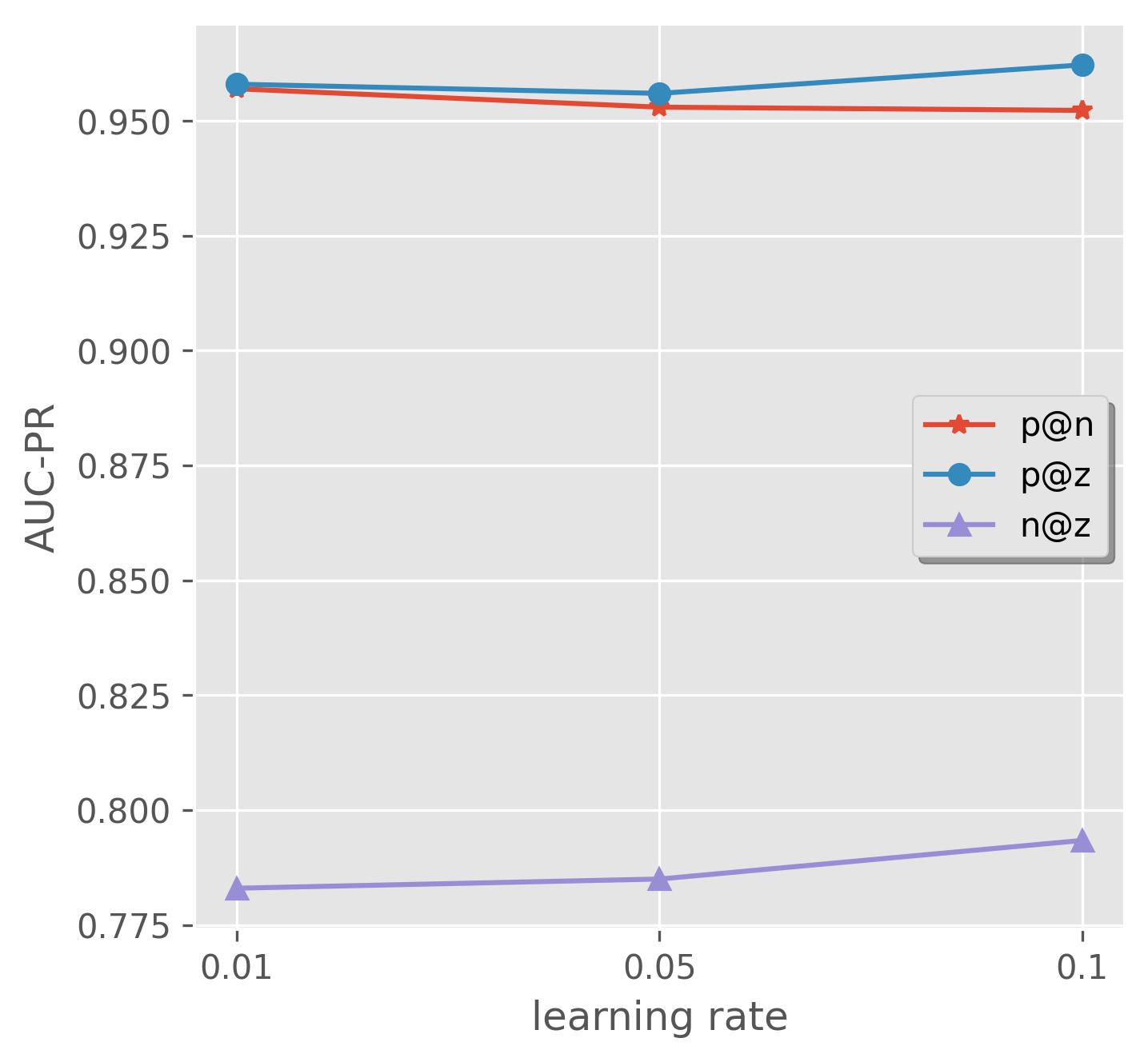}
        \caption{}
\end{subfigure}
\caption{\textsl{wikiElec}: Performance of \textsc{SLIM} across learning rates for different tasks, (a) Area-Under-Curve Receiver Operating Characteristic scores, (b) Area-Under-Curve Precision-Recall scores. Both AUC-ROC and AUC-PR scores are constant across different dimensions }
\label{fig:lr_roc_pr1}
\end{figure*}

The learning rate for \textsc{SLDM} and \textsc{SLIM} was set to $lr=0.05$. In Fig \ref{fig:lr_roc_pr1}, we provide the performance across different tasks for the \textsl{wikiElec} dataset, considering three different learning rates $lr\in\{0.01,0.05,0.1\}$. We observe that the performance can be considered constant for the different learning rates, showing small sensitivity to the choice of this hyperparameter.

\clearpage

% \clearpage
% \bibliography{reference}

% \end{document}